\documentclass[pdflatex,sn-mathphys-ay]{sn-jnl}

\usepackage{graphicx}%
\usepackage{multirow}%
\usepackage{amsmath,amssymb,amsfonts}%
\usepackage{amsthm}%
\usepackage{mathrsfs}%
\usepackage[title]{appendix}%
\usepackage{xcolor}%
\usepackage{textcomp}%
\usepackage{manyfoot}%
\usepackage{booktabs}%
\usepackage{algorithm}%
\usepackage{algorithmicx}%
\usepackage{algpseudocode}%
\usepackage{listings}%

\usepackage[T1]{fontenc}
\usepackage{graphicx}
\usepackage{caption}
\usepackage{subcaption}
\usepackage{float}
\usepackage{dirtytalk}
\usepackage{tabularx}
\usepackage{orcidlink}
\usepackage{placeins}

\begin{document}
 
\title[Predicting potentially abusive clauses in Chilean terms of services]{Predicting potentially abusive clauses in Chilean terms of services with natural language processing}

\author*[1]{\fnm{Christoffer} \sur{L\"offler}\orcidlink{0000-0003-1834-8323}}\email{christoffer.loffler@pucv.cl}
\author[2]{\fnm{Andrea} \sur{Martínez Freile}\orcidlink{0000-0003-1275-2763}}\email{andrea.martinez@uai.cl}
\author[2]{\fnm{Tomás} \sur{Rey Pizarro}\orcidlink{0009-0006-5753-1680}}\email{torey@alumnos.uai.cl}

\affil*[1]{\orgdiv{School of Computer Engineering}, \orgname{Pontificia Universidad Católica de Valparaíso}, \orgaddress{\street{Brasil 2950}, \city{Valparaíso}, \postcode{2340025}, \country{Chile}}}

\affil[2]{\orgdiv{Faculty of Law}, \orgname{Universidad Adolfo Ibáñez}, \orgaddress{\street{Av. Padre Hurtado 750}, \city{Viña del Mar}, \postcode{2581793}, \country{Chile}}}


\abstract{
This study addresses the growing concern about the inclusion of abusive clauses in consumer contracts, exacerbated by the proliferation of online services with complex Terms of Service that are rarely read. Even though research on automatic analysis methods is conducted, the difficulty of detecting such clauses is aggravated by the general focus on English-language Machine Learning approaches and on major jurisdictions, such as the European Union.
We introduce a new methodology and a substantial Spanish-language dataset addressing this gap. We propose a novel annotation scheme with four categories and 20 classes and apply it to 50 online Terms of Service used in Chile. Our evaluation of transformer-based models highlights how factors like language- and/or domain-specific pre-training, few-shot sample size, and model architecture affect the detection and classification of potentially abusive clauses. 
Results show a large variability in performance for the different tasks and models, with the highest macro-F1 scores for the detection task ranging from 79\% to 89\% and micro-F1 scores up to 96\%, while macro-F1 scores for the classification task range from 60\% to 70\% and micro-F1 scores from 64\% to 80\%. 
Notably, this is the first Spanish-language multi-label classification dataset for legal clauses, applying Chilean law and offering a comprehensive evaluation of Spanish-language models in the legal domain. 
Our work lays the ground for future research in method development for rarely considered legal analysis and potentially leads to practical applications to support consumers in Chile and Latin America as a whole.
}


\maketitle

\section{Introduction}\label{sec:intro}


In the digital age, where a growing number of products and services are offered through online platforms, Terms of Service (ToS) have become increasingly relevant, since they usually constitute the step where the provider and the user become legally bound. However, ToS are long, difficult to read, sometimes not even available in the user's mother tongue, and at the end of the day, pointless to review. Even if one could and would want to spend time reading them to use the service, there is no alternative but to accept them. It is impossible to change or negotiate a disadvantageous or even an abusive clause~\citep{delamazagazmuriContratosPorAdhesion2003}. These problems of rational apathy and information asymmetry are present in most modern consumer contracts, but now they are amplified by the number and speed they occur through digital platforms. This constitutes a challenge for the regulator, the surveillance agencies, and the consumers~\citep{lippiCLAUDETTEAutomatedDetector2019}.

In this context, the use of Artificial Intelligence in the legal field could be of tremendous impact, because it could automate the revision of these ToS and other consumer contracts, alerting the user and national services of the existence of potentially abusive clauses~\citep{lippiForceAwakensArtificial2020}.
Recent developments have already led to promising systems, however, with few exceptions~\citep{tocchiniDetectionPotentiallyNoncompliant2024}, these have mainly focused on the European Union~\citep{lippiCLAUDETTEAutomatedDetector2019,braun2021automated,drawzeskiCorpusMultilingualAnalysis2021,ruggeriDetectingExplainingUnfairness2022,braun-matthes-2024-agb,dadasSupportSystemDetection2024,galassiUnfairClauseDetection2024}. Methods developed for other legal systems and languages, such as South American legislation developed in Spanish, are also necessary and should be included in the global research community’s attention.  

Hence, our research objectives focus on developing Machine Learning (ML) for the automated identification of potentially abusive clauses in Chilean online ToS. Hereby, we build on work conducted on European online ToS~\citep{lippiCLAUDETTEAutomatedDetector2019, ruggeriDetectingExplainingUnfairness2022, chalkidis-etal-2022-lexglue,dadasSupportSystemDetection2024,galassiUnfairClauseDetection2024} and update the methodology to recent developments in Natural Language Processing (NLP)~\citep{openaiGPT4TechnicalReport2024,dubeyLlamaHerdModels2024,yangQwen2TechnicalReport2024,mistralaiteamAIAbundance2024}.

Thus, our contributions are as follows:
\begin{itemize}
    \item We present a classification system categorizing 20 potentially abusive clauses into four groups of which we annotate three for machine learning tasks: 6 illegal, 6 dark, and 8 gray clauses. 
    \item We introduce the Chilean ToS dataset, comprising 50 legal documents in Spanish with a total of 5,209 annotated clauses. From these, we generate six datasets for detection and classification tasks. These datasets are imbalanced and can be used for binary or multi-class/-label classification.
    \item We conduct a comprehensive fine-tuning study of state-of-the-art Large Language Models (LLMs). We compare the baselines, including a linear Support Vector Machine (SVM)~\citep{cortesSupportvectorNetworks1995} with TF-IDF features, with English-language pre-trained models such as BERT~\citep{devlinBERTPretrainingDeep2019}, RoBERTa~\citep{liuRoBERTaRobustlyOptimized2019}, DeBERTav3~\citep{he2023debertav}, and LongFormer~\citep{beltagyLongformerLongDocumentTransformer2020}. We also include multilingual models like M-BERT~\citep{devlinBERTPretrainingDeep2019}, XLM-RoBERTa~\citep{conneau-etal-2020-unsupervised}, domain-specific models such as Legal-BERT~\citep{chalkidis-etal-2020-legal}, and Spanish-language pre-trained models like BETO and DistilBETO~\citep{caneteSpanishPretrainedBERT2023}. Notably, to the best of our knowledge, we are the first to evaluate the Spanish-language legal-domain pre-trained RoBERTalex~\citep{gutierrez-fandinoSpanishLegaleseLanguage2021} on a Spanish-language legal task.
    \item We explore the few-shot capabilities of multilingual LLMs and compare their performance with the results obtained from fine-tuned models. The study includes the commercial models GPT-4o/4o-mini~\citep{openaiGPT4TechnicalReport2024} and the open models Qwen2.5~\citep{yangQwen2TechnicalReport2024}, Llama3.1~\citep{dubeyLlamaHerdModels2024}, Gemma2~\citep{gemmateamGemma2Improving2024}, and Mistral-small~\citep{mistralaiteamAIAbundance2024} and -nemo~\citep{mistralaiteamMistralNeMo2024}. 
\end{itemize}

 
The rest of the paper is structured as follows. The next section formulates the problem from the Chilean legal perspective. Section~\ref{sec:chile} presents the proposed annotation system and introduces the Chilean Terms of Services corpus and its datasets. Section~\ref{sec:method} explains ML methods and Section~\ref{sec:results} describes the experimental setup and reports the results of our study. Section~\ref{sec:discussion} discusses relevant prior research and provides an outlook, and Section~\ref{sec:conclusion} concludes.

\section{Problem Formulation}\label{sec:problem}

In contracts of adhesion, that is, contracts, where one party sets the terms and conditions of the agreement and the counterpart is generally left only with the faculty to accept or reject the proposal as a whole~\citep{delamazagazmuriContratosPorAdhesion2003}, there is the risk that abusive clauses would be included. Legal systems refer to these types of clauses in various ways. In Chile, the Consumer Protection Law Act (Law Number 19,496, "Ley de Protección al Consumidor", recently updated and modified by the "Decreto con Fuerza de Ley Numero 3" from 2021, from now on LPC) does not define abusive clauses. Instead, it outlines the requirements that every consumer contract must fulfill and lists clauses that will be deemed invalid if introduced and that could potentially nullify the entire contract (Article 16 of the LPC).

The Chilean National Service for Consumer Protection ("Servicio Nacional del Consumidor", SERNAC) has defined abusive clauses as those imposed by the provider of the goods or services in violation of the principles of equality and good faith, that result in a significant imbalance of contractual duties, in detriment of the weakest part, the consumer~\citep{sernacResolucionExentaNdeg9312021}. Examples of these clauses are the ones that allow the unilateral termination of the contract at the mere discretion of the provider, that is, without expression of cause or the fault of the consumer, that include disclaimers of liability even when the suppliers caused the damage, or that cause a reversal of the burden of proof, among many other abusive clauses.  

The core of these types of clauses lies in the power imbalance between the parties involved in the contract. One way to restore that balance is to report the existence of abusive clauses and eventually pursue legal action against providers who use them. However, the first requirement for this process is that consumers are aware of the existence of abusive clauses, which is not a simple task.

According to a survey conducted by SERNAC, over half of consumers are unable to identify potentially abusive clauses in contracts, while a third are completely unaware of their existence.\footnote{\url{https://www.sernac.cl/portal/604/w3-article-64918.html}, (March 1st, 2022)\label{footnote:sernac:study}} This situation is exacerbated even further in the case of ToS, which are designed to be accepted just by one \textit{click} in a matter of seconds~\citep{lippiCLAUDETTEAutomatedDetector2019}. 

In this context, the general public could use specialized ML methods as a preventive tool to check if a ToS or other contracts include potentially abusive clauses. The model could swiftly accomplish the task, that would be arduous and time-consuming for a human. This model could automatically review the entire extension of the terms and raise an alarm if detects any clause that appears to be abusive. The primary objective of this research is to develop such methods.

It is crucial to recognize that specific clauses can be interpreted in various ways, depending on the systematic analysis of the contract and the factual context in which they are applied. This causes interpretation problems that affect even specialized lawyers, posing a substantial challenge for ML algorithms. Furthermore, the judicial authority determines whether a clause is abusive. They will impose the appropriate consequences, which may include the annulment of the clause or contract, and eventually, the compensation of the suffered damages. Until the Supreme Court has settled the discussion, a clause is only potentially abusive. In other words, neither an AI method, lawyer nor even SERNAC can definitively determine whether a clause is abusive. Only the Supreme Court possesses the authority to affirm or deny a clause's nature with legal certainty. Thus, our proposed ML methods and their results should be understood as a reference to guide further actions but are not intended to replace professional and technical advice. However, they can still serve as a very useful preventive tool that can raise awareness among consumers about the existence of abusive clauses and may support the work of surveillance agencies and lawyers who deal with such clauses.  

\section{Potentially Abusive Clauses}\label{sec:chile}

In this study, we analyzed 50 Spanish-language ToS from companies that provide services in Chile.\footnote{Academia.edu, Airbnb, Amazon, App Copec, Apple, Badoo, Bluexpress, Booking, Box, Canva, Despegar, Dropbox, eBay, Edreams, Evernote, Facebook, Fitbit, Google, Headspace, Instagram, Linkedln, Mercadolibre, Microsoft, MyHeritage, Nespresso, Netflix, Nike, Nintendo, Paris, Pokemon GO, Pullman Bus, Rappi, Ripley, Rovio, Skype, Skyscanner, Snapchat, Spotify, Starbucks, Tenpo, Tinder, TripAdvisor, Uber, Vimeo, WhatsApp, Wild Foods, World of Warcraft, X (Twitter), Yahoo, YouTube}  In total, this corresponds to 5,209 clauses, of which $20.3\%$ were identified as potentially abusive under at least one category. That means they were clauses that presented more than one reason that deemed them abusive, so the level of abusiveness is actually underrepresented. That is a surprisingly high level of abusive clauses since one could expect that standardized contracts used by internationally renowned corporations should comply with the law. Again this result reaffirms the necessity of developing tools that alert consumers of the existence of potentially abusive clauses. 

From the about 20\% of clauses that were deemed abusive by our legal expert annotators, some of them were quite common. For example, we found 156 clauses in the corpus that limit the provider's liability (ltd). On the other hand, there are extremely rare cases, such as the 10 clauses in the corpus that permit the provider the unilateral termination of the contract (ter). Furthermore, several clauses may be potentially abusive in multiple ways. For instance, out of the 117 clauses that impose external costs on the consumer (des risk), 34 were also against good faith (bfe).  In the following sections, we describe our proposed categorization system and its type of potentially abusive clauses, the annotated clauses and finally the corpus statistics.

\subsection{Categorization System}\label{sec:system}

We developed our own categorization system for reviewing the ToS based on three factors. First, we used the legal classification Chilean doctrine has made of Abusive Clauses~\citep{delamazagazmuriContratosPorAdhesion2003,lopezdiazHyperVulnerableConsumer2022,barrientoscamusLeccionesDerechoConsumidor2019,moralesAlgunosProblemasExtension2018,pizarrowilsonFRACASOSISTEMAANALISIS2007,sernacResolucionExentaNdeg9312021}. Second, we reviewed the classification used in previous related work~\citep{lippiCLAUDETTEAutomatedDetector2019,ruggeriDetectingExplainingUnfairness2022}. Finally, we took the optimal methodology to identify the clauses into consideration. 
This work defines four categories of potentially abusive clauses for Chilean law: clauses that infringe formal requirements, and illegal, dark, or gray clauses. 
Only the last three of these four are predicted in our experiments, as the next section will clarify.

\subsection{Formal Requirements of Clauses}\label{sec:formal requirements of clauses}

Some norms of the LPC, like Articles 17 and 17A, specify different formal requirements that contracts must fulfill regardless of their type or content. For example, the minimum size of the letter that should be used, the language in which they should be written, and the obligation of providing a print copy of the contract, among others. Under these requirements, many of the ToS were at fault for very basic elements, for example, some of them were not available in Spanish\footnote{Those were excluded from this study that focus on this language}, or were not printable. 

Even though relevant to consumers, these types of infringements of the law are not of special interest in our current research, since their identification is relatively straightforward and no specialized model is required for it. Instead, more traditional text parsers could be used to analyze, for example, the language of the ToS, its HTML formatting and font size, or the blocking of any print functionality via JavaScript. Therefore, they were excluded from the experimental analysis of this work, which focuses on the development and review of the performance of ML methods.

\subsection{Illegal Clauses}\label{sec:illegal}

Illegal Clauses refer to provisions that go against the explicit order of a norm to the detriment of the consumer. Therefore, even though they are included in a contract or ToS, they are not effective and should be considered as "not written", according to Articles 1,467 and 1,682 of the Civil Code ("Código Civil", CC) regarding the effect of annulment of clauses. 

This category is not used by the Chilean doctrine or previous related work but is of our creation. We decided to separate this type of clause due to two reasons. First, because they constitute a relevant case of potentially abusive clauses (about 45\% of them), and, secondly, because their characteristics make them both easy to recognize and simultaneously uniquely challenging for an AI system. They are easily recognizable because they are contrasted directly with a written norm that forbids them. However, the condition for their detection is the awareness of the legal rule being infringed, which constitutes the challenge of the category. The difficulty arises from the broad scope of potentially applicable norms because it is challenging to cover all the relevant ones since a contractual provision could potentially violate any legal rule of the Chilean system. Even though the ML model may successfully predict the lawfulness of clauses in an annotated dataset, the model is ultimately limited by its training scope. Therefore, it can be challenging to identify and group the applicable norms. For example, during the annotation of ToS, we found that a specific law was infringed by a ToS, that is, Law Number 20,720 regarding the bankruptcy regime and insolvency procedures. This was possible due to the annotators' expert knowledge of bankruptcy law, but could not have been generally expected from a lawyer specialized in other fields. Similarly, maybe other clauses were unlawful but were not detected by the annotator for being outside of their scope of knowledge, and thus the ML model was not trained with the required data. 

In the process, we noticed that some illegalities were more common than others, so we split them into subcategories. For example, the clauses that contravene provisions of the contractual liability regime of the CC included infringements in the rules for the determination of damages or the binding nature of contracts. Other usual illegal clauses were the ones that altered the jurisdiction of courts or prevented the consumer from acceding to one. Instead, they forced them to use informal conflict mechanism resolutions, violating procedural rules of the LPC, the Civil Procedure Code ("Código de Procedimiento Civil", CPC), and the Organic Code of Courts ("Código Orgánico de Tribunales, COT). Since many ToS come from international corporations, clauses referencing foreign law, that are not applicable in Chile (and sometimes even in direct conflict with national provisions), were also identified as illegal, to ensure local consumer rights are upheld.

\subsection{Dark Clauses}\label{sec:dark}

Dark Clauses are the ones that, even though they are not directly illegal, are manifestly abusive. They are caused by the great power imbalance the provider holds, since the consumer can only agree with them or not sign the contract, i.e., not use the service. The provisions included in letters a), b), c) d), e) and h) of Article 16 LPC refer to this type of clauses~\citep{barrientoscamusLeccionesDerechoConsumidor2019}. These are, for example, the unilateral termination of the contract at the mere discretion of the provider; the arbitrary modification of the elements of the contract, including a unilateral price increase; preemptive limitation of liability clauses; and alteration of the burden of proof. 

A commonality among these clauses is how evidently imbalanced they are, and for that reason, the Chilean legislator presumes their abusiveness. This means that, even though their identification as abusive still needs to be done by a court, the burden of proof lies with the supplier and not on the consumer. In other words, the provider that incorporated a dark clause into their contract needs to prove that, despite their appearance, they are not actually abusive. This constitutes an important advantage at the moment of presenting a claim, but a requirement for this step is still that consumers are aware of their existence and abusiveness. According to the studies conducted by SERNAC (see footnote~\ref{footnote:sernac:study}), this is not a simple task for most people. Perhaps it is because of this that, despite the legislator identifying them as abusive, they continue to be widely used by providers.

Dark Clauses present three special characteristics that might be beneficial for the use of ML. First, the above-mentioned letters of Article 16 LPC constitute a well-defined list of clauses, that stands in contrast with the countless sources of unlawfulness that can make a clause illegal. Closed-set classification problems, such as this, are generally considered easier to solve than open-set classification like the Illegal Clauses problem. For instance, there is no need for open-set recognition~\citep{vaze2022openset}. However, the challenge of the variability of examples for one type of potentially abusive clause remains, i.e., they can be reformulated in different ways that are not recognizable anymore, for instance, using synonyms or a different grammatical structure.
Second, since they are presumed abusive, there are fewer errors of interpretation and they are relatively clear. In contrast, Gray Clauses are more ambiguous, see Sec.~\ref{sec:gray}. While the final decision is with the courts, their identification via ML would help raise awareness of the clauses' existence among consumers. Third, since many Dark Clauses appear relatively frequently in the ToS that we reviewed, and thus, their patterns and specific formulations repeat as well, ML models may learn to predict these more easily. Conversely, the clauses that are less represented in the data may be harder to predict accurately.

\subsection{Gray Clauses}\label{sec:gray}

Gray Clauses are provisions that, in contrast with the Dark Clauses, are not easily identifiable as abusive. Their legal base is Article 16 letter g) LPC, which refers, in general, to clauses that are contrary to "good faith". That is, they cause a significant imbalance in the rights and obligations of the parties of the contract to the detriment of the consumers. This definition is broad and overlaps with the concept of the abusive clause itself. It also does not provide clear guidelines to proceed. On the contrary, it is meant to constitute an open category to include any potentially abusive clauses not accounted for by the legislator in the previous cases of the norm. The specific situations comprised under this umbrella provision are a matter of interpretation. Due to this, there is no presumption of abusiveness. That means, that the consumer will have to prove in court that the disputed clause is abusive under the terms of Article 16 letter g) LPC~\citep{barrientoscamusLeccionesDerechoConsumidor2019}. 

The ambiguity of this category makes it the hardest of all the ones analyzed, both for the consumers, the legal experts, and also for ML models. Their identification will usually depend on the interconnected understanding of different sections of the ToS and even of contextual facts not present in the wording of the documents. Furthermore, their determination will also depend on the legal argumentation of the plaintiff and defendants, and finally, on the opinion of the court. 

However, during the analysis of the ToS, our legal experts were able to identify eight different classes of Gray Clauses, based on the contractual imbalance that they generate, and on the frequency that they were found. Examples of these clauses are the ones that impose risks or costs on the consumer arising from external factors outside of the contractual relationship, such as \textit{force majeur}\footnote{This refers to unforeseeable circumstances that are out of the control of the parties of a contract, for example, natural disasters}; that grant the provider the ability to share information of the consumer with unrelated parties, or that grant the provider broad, perpetual, or irrevocable faculties that extend beyond the contract's duration; among other clauses. These classes of Gray Clauses are suggestions of our legal experts but could be extended by analyzing other types of consumer contracts and ToS, for example, financial services. 

Whether a clause is defined as Dark or Gray may also be a matter of degree that depends on the legal argumentation of the reviewer. For example, Article 16 letter a) LPC clearly states that clauses that allow the modification of contracts at the sole discretion of the provider or allow its unilateral termination are not permitted. These would be considered Dark Clauses under our categorization, as was the case when we found clauses that allow changes or terminations of services without previous notice during the revision of the ToS. However, when a change was notified with enough anticipation, for example, a month in advance, the legal experts considered that the "unilateral" nature of the change was altered, since the consumer may have the opportunity to be aware of the future change and possibly negotiate the modifications or its consequences. This might be only potentially feasible since the consumer might still not have any bargaining power to prevent the change and the only alternative remaining might be to leave the service. The clause shows an imbalance between the parties to the detriment of the consumer and therefore was labeled as a Gray Clause, but since the outcome is equivalent to the first example, this one could also be considered a Dark Clause by a stricter interpreter. 

For the annotations and the learning process of the model, we considered it very valuable to also distinguish these slightly different variations of similar circumstances. As in the example of unilateral modification, a model trained on samples without previous notification by the provider may have low precision once the grammatical structure of the sample is changed to include contractual changes with previous notifications.
For consumers, it is valuable to highlight different potentially abusive practices to raise awareness of these situations.   

\subsection{Annotated Clauses}\label{sec:clauses}

From the four categories of Abusive Clauses identified in our research, we analyze three with our ML methodology: the Illegal, the Dark, and the Gray Clauses. For these categories, we select 20 classes according to the legal source used to define them and the frequency they appear in ToS. This section describes the annotated clauses in detail, introducing their keys in the dataset, their legal sources, and their explanations in the following Table~\ref{tab:annotated_clauses}.

\begin{table*}[t]
    \centering
    \resizebox{0.95\textwidth}{!}{
        \begin{tabular}{|p{1.1cm}|p{10.9cm}|p{2.3cm}|}
            \hline
            \multicolumn{3}{|c|}{\textbf{Illegal Clauses}} \\
            \hline
            \textbf{Key} & \textbf{Explanation} & \textbf{Legal Source}  \\
            \hline
            lpc pro & Clauses that violate the prohibition of altering jurisdiction rules of the LPC & Art. 50-A and 50-H LPC \\
            \hline
            lpc int & Clauses that modify the liability of intermediaries of goods or services & Art. 43 LPC \\
            \hline
            lpc jus & Clauses that forbid access to justice in a broader sense& Art. 50 H ff. LPC, Art. 253 ff. CPC, Art. 108 ff. COT\\
            \hline
            lpc & Clauses that violate a provision of the LPC that does not refer to the list of abusive clauses of Article 16. This constitutes a residual class & LPC\\
            \hline
            cc rc & Clauses that violate a provision of the Civil Liability Regime of the Civil Code & Art. 1489, 1545, 1556 CC\\
            \hline
            na & Clauses that mention a foreign law, which is not applicable in Chile & Art. 16 CC\\
            \hline
            \hline
            \multicolumn{3}{|c|}{\textbf{Dark Clauses}} \\
            \hline
            \textbf{Key} & \textbf{Explanation} & \textbf{Legal Source}  \\
            \hline
            cr & Clauses that include a unilateral and arbitrary modification of the ToS without the consumer being able to challenge or reject the change & Art. 16 a) LPC\\
            \hline
            ter & Clauses that include a unilateral and arbitrary termination of services & Art. 16 a) LPC\\
            \hline
            ch & Clauses that allow unilateral modifications on the price of the service & Art. 16 b) LPC\\
            \hline
            er & Clauses that let the consumer bear the consequences of administrative or internal errors of the provider & Art. 16 c) LPC\\
            \hline
            ltd & Clauses that include limitations on the liability of the provider & Art. 16 e) LPC\\
            \hline
            nod & Clauses that substantially limit the way consumers can exercise their rights & Art. 16 h) LPC\\
            \hline
            \hline
            \multicolumn{3}{|c|}{\textbf{Gray Clauses}} \\
            \hline
            \textbf{Key} & \textbf{Explanation} & \textbf{Legal Source}  \\
            \hline
            des risk & Clauses that impose risks or costs on the consumer that are unrelated to the provider arising from external factors to the contractual relationship, such as \textit{force majeure} & Art. 16 g) LPC\\
            \hline
            des def &  Clauses that impose obligations on the consumer to defend, indemnify, or cooperate in a potential litigation in favor of the provider & Art. 16 g) LPC\\
            \hline
            des inf & Clauses that allow the provider to share information of the consumer with third parties unrelated to the service & Art. 16 g) LPC\\
            \hline
            des lic & Clauses granting the provider broad and excessive faculties that can also extend after the termination of the service & Art. 16 g) LPC\\
            \hline
            des det & Clauses that allow the provider to resolve conflicts with the consumer through internal processes, potentially affecting consumer rights & Art. 16 g) LPC \\
            \hline
            des reser & Clauses that allow the modification of the contract at the discretion of the provider, but with a lesser level of "unilaterality" than Dark Clauses & Art. 16 g) LPC\\
            \hline
            des uni & Clauses that allow the modification of the contract with previous notice, changing the level of "unilaterality" required for Dark Clauses & Art. 16 g) LPC\\
            \hline
            bfe & Clauses that go against the requirements of good faith taking into consideration the purpose of the contract. This constitutes a residual class & Art. 16 g) LPC\\
            \hline
        \end{tabular}
    }
    \caption{The annotation scheme consists of the potentially abusive clauses' keys in the dataset, their legal sources, and their explanations}
    \label{tab:annotated_clauses}
\end{table*}

\subsection{Corpus}\label{sec:stats}

\begin{table}[t]
\centering
\begin{minipage}{0.32\columnwidth}
    \centering
    \begin{tabular}{|p{1cm}||p{1cm}|p{1cm}|p{1cm}|p{1cm}|}
    \hline
    \multicolumn{5}{|c|}{\textbf{Illegal Detection}} \\
    \hline
     & Train  & Val & Test & Sum \\
    \hline
    ok & 2040 & 875 & 1250 & 4165 \\
    abusive & 295 & 127 & 181 & 603 \\
    \hline
    \textbf{Sum} & \textbf{2335} & \textbf{1002} & \textbf{1432} & \textbf{4768} \\
    \hline
    \end{tabular}
\end{minipage}
\hfill
\begin{minipage}{0.32\columnwidth}
    \centering
    \begin{tabular}{|p{1cm}||p{1cm}|p{1cm}|p{1cm}|p{1cm}|}
    \hline
    \multicolumn{5}{|c|}{\textbf{Dark Detection}} \\
    \hline
     & Train & Val & Test & Sum \\
    \hline
    ok & 2040 & 875 & 1250 & 4165 \\
    abusive & 170 & 74 & 105 & 349 \\
    \hline
    \textbf{Sum} & \textbf{2210} & \textbf{949} & \textbf{1355} & \textbf{4514} \\
    \hline
    \end{tabular}
\end{minipage}
\hfill
\begin{minipage}{0.32\columnwidth}
    \centering
    \begin{tabular}{|p{1cm}||p{1cm}|p{1cm}|p{1cm}|p{1cm}|}
    \hline
    \multicolumn{5}{|c|}{\textbf{Gray Detection}} \\
    \hline
     & Train & Val & Test & Sum \\
    \hline
    ok & 2040 & 875 & 1250 & 4165 \\
    abusive & 185 & 79 & 113 & 377 \\
    \hline
    \textbf{Sum} & \textbf{2225} & \textbf{954} & \textbf{1363} & \textbf{4542} \\
    \hline
    \end{tabular}
\end{minipage}
\caption{Dataset splits for detection using iterative stratified sampling.}\label{tab:detect_stats}
\end{table}

\begin{table}[t]
\centering
\begin{minipage}[t]{0.32\textwidth}
    \centering
    \begin{tabular}{|p{1.2cm}||p{1cm}|p{1cm}|p{1cm}|p{1cm}|}
    \hline
    \multicolumn{5}{|c|}{\textbf{Illegal Classification}} \\
    \hline
    Label & Train & Val & Test & Sum \\
    \hline
    na & 140 & 60 & 86 & 286 \\
    lpc pro & 67 & 29 & 41 & 137 \\
    cc rc & 27 & 12 & 17 & 56 \\
    lpc & 22 & 9 & 13 & 44 \\
    lpc int & 20 & 9 & 12 & 41 \\
    lpc jus & 19 & 8 & 12 & 39 \\
    \hline
    \textbf{Sum} & \textbf{335} & \textbf{127} & \textbf{181} & \textbf{643} \\
    \hline
    \end{tabular}
\end{minipage}
\hfill
\begin{minipage}[t]{0.32\textwidth}
    \centering
    \begin{tabular}{|p{1.2cm}||p{1cm}|p{1cm}|p{1cm}|p{1cm}|}
    \hline
    \multicolumn{5}{|c|}{\textbf{Dark Classification}} \\
    \hline
    Label & Train & Val & Test & Sum \\
    \hline
    ltd & 76 & 33 & 47 & 156 \\
    cr & 48 & 21 & 29 & 98 \\
    nod & 19 & 8 & 12 & 39 \\
    er & 17 & 8 & 11 & 36 \\
    ch & 5 & 2 & 3 & 10 \\
    ter & 5 & 2 & 3 & 10 \\
    \hline
    \textbf{Sum} & \textbf{170} & \textbf{75} & \textbf{105} & \textbf{349} \\
    \hline
    \end{tabular}
\end{minipage}
\hfill
\begin{minipage}[t]{0.32\textwidth}
    \centering
    \begin{tabular}{|p{1.2cm}||p{1cm}|p{1cm}|p{1cm}|p{1cm}|}
    \hline
    \multicolumn{5}{|c|}{\textbf{Gray Classification}} \\
    \hline
    Label & Train & Val & Test & Sum \\
    \hline
    bfe & 37 & 16 & 23 & 83 \\
    des risk & 41 & 17 & 25 & 83 \\
    des reser & 29 & 12 & 17 & 58 \\
    des uni & 24 & 10 & 14 & 48 \\
    des det & 17 & 8 & 11 & 36 \\
    des def & 17 & 7 & 10 & 34 \\
    des inf & 12 & 5 & 8 & 25 \\
    des lic & 8 & 4 & 5 & 17 \\
    \hline
    \textbf{Sum} & \textbf{185} & \textbf{79} & \textbf{113} & \textbf{384} \\
    \hline
    \end{tabular}
\end{minipage}
\caption{Dataset splits for multi-class classification using iterative stratified sampling.}\label{tab:classify_stats}
\end{table}
The 50 ToS analyzed comprise a total of 6,523 clauses.
If there were multiple versions of the ToS available, we selected the most recent one for Chilean customers.

The annotations were performed by two Chilean legal experts\footnote{A licensed lawyer and professor with PhD in Commercial Law and a graduate with a Bachelor of Law undergoing mandatory state practice to obtain their professional license from the Chilean Supreme Court.}, who we recruited in-house. We used a three-step process to ensure quality control of the annotation process by expert review~\citep{monarchHumanLoopMachineLearning2021}. The first expert fully annotated a contract, and then the second expert reviewed and commented on the annotations. Ultimately, both experts discussed and agreed on the final labels. Due to the expert review, we do not report agreement measures such as Cohen's kappa~\citep{monarchHumanLoopMachineLearning2021}.

The experts first segmented the text along its paragraphs into clauses. Then, they applied the annotation scheme on the clauses and marked them with one or more keys, refer to Table~\ref{tab:annotated_clauses} for details. Of the 20\% that are identified as abusive, around 45\% are categorized as illegal, 26\% as dark, and 28\% as gray. For a detailed breakdown, refer to Table~\ref{tab:stats_total}. 
Following Section~\ref{sec:clauses}, we aggregate and recombine the annotated clauses into the datasets Illegal, Dark, and Gray Clauses. For each category, we define the binary classification task of detection and the multi-label multi-class task of classification. We eliminate any clause that comprises fewer than seven words. This yields a total corpus size of 5,209 annotated clauses that we use to construct six datasets.

For each detection task, we select one of the three categories of potentially abusive clauses and reduce its labels to the uniform label "abusive". Then, we combine it with the non-abusive clauses, see Table.~\ref{tab:detect_stats} for an overview. This pre-processing step emphasizes each category's unique challenges and creates isolated detection tasks. This is especially interesting for the novel category of Illegal Clauses, which has not been studied previously. For future work, we envision creating one model that detects the combined categories at once.

\begin{figure}[b]
    \centering
    \includegraphics[width=0.99\textwidth]{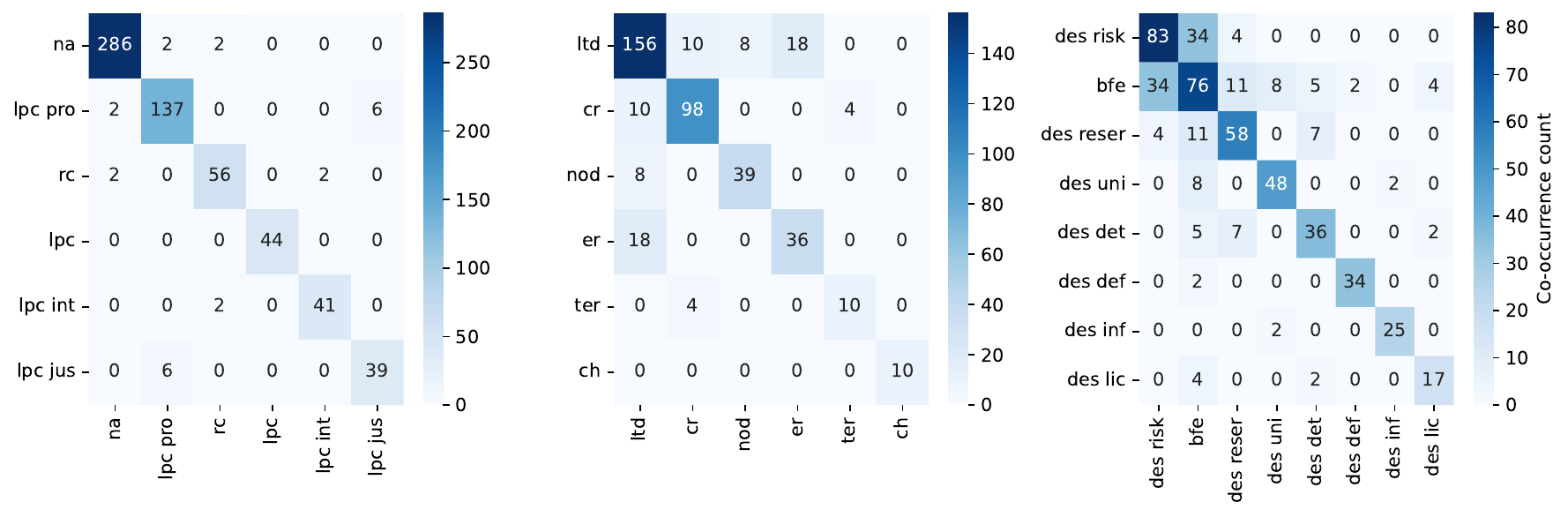}
    \caption{The co-occurrence of classes for the multi-label task Illegal Classification (left) exhibits a very low co-occurrence between classes. In contrast, the harder datasets Dark Classification (middle) and Gray Classification (right) exhibit a high rate of co-occurrence for specific label combinations}
    \label{fig:co-occurence}
\end{figure}

The classification task then consists of only the paragraphs that are labeled potentially abusive, see Table.~\ref{tab:classify_stats} for an overview. These tasks are inherently multi-label, and it is common to find combinations of labels for a single clause, see the co-occurrence in Fig.~\ref{fig:co-occurence}. 
However, our corpus exhibits varying levels of co-occurrence, ranging from the almost pure annotations in Illegal Classification to the high rates of co-occurrence in Dark Classification and particularly in Gray Classification.

Since some combinations of multi-label annotations are extremely rare, we employed an iterative stratified sampling method~\citep{sechidisStratificationMultilabelData2011} to account for the significant disparity in the number of distinct label sets. This method ensures a highly representative split into training, validation, and testing, given the complexity of the annotations. Our use of this method involves dividing each subset into training and testing sets by a ratio of 70\% to 30\%, and then further subdividing the training set into training and validation sets, with a ratio of 70\% to 30\%. 

The iterative stratified sampling technique~\citep{sechidisStratificationMultilabelData2011} for the multi-label data causes the total sum of instances to be different to that of the corresponding detection task to sample a balanced split. For example, the algorithm selectively over-samples to stratify the Illegal Classification dataset split to a total of 643 samples instead of 603 in the detection task.

\begin{figure}[h]
    \centering
    \includegraphics[width=0.9\textwidth]{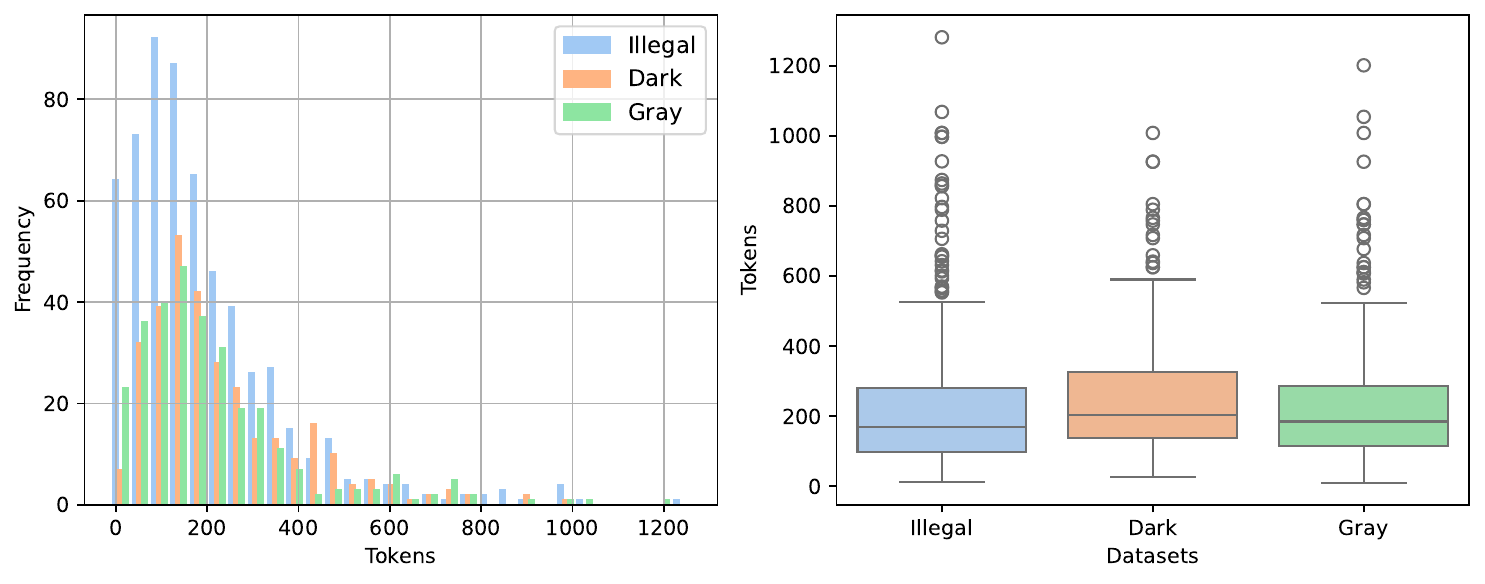}
    \caption{We calculate token statistics for the classification tasks using the GPT2 tokenizer~\citep{radford2019language} and present each task's statistics as histogram (left) and as box plot (right)}
    \label{fig:token_stats}
\end{figure}

The average sequence length of the entire dataset using the GPT2 tokenizer~\citep{radford2019language} is 230.92 tokens (median 181 tokens). Furthermore, $90\%$ of the clauses consist of fewer than 461 tokens, which is a relevant feature for the model selection, as models' sequence length can be a limiting factor. Respective histograms and box plots for the datasets' token statistics are provided in Fig.~\ref{fig:token_stats}.

\section{Methodology}\label{sec:method}

We employ two learning paradigms to detect and classify potentially abusive clauses. First, we fine-tune transformer-based models. Second, we evaluate the few-shot performance of commercial and open LLMs. We follow \cite{chalkidis-etal-2022-lexglue} and include a linear Support Vector Machine~\citep{cortesSupportvectorNetworks1995} as a baseline. This SVM uses TF-IDF features for the top-$K$ frequent $n$-grams ($n\in[1,2,3]$). 

\subsection{Fine-Tuning Classifier Models}\label{sec:finetune}

In this study, we experiment with pre-trained transformer models~\citep{10.5555/3295222.3295349}. \cite{grecoBringingOrderRealm2024} conduct a systematic review that delves deeper into the various models commonly employed for legal tasks. We employ models that were either pre-trained solely on English or Spanish, or on multiple languages simultaneously, or on English- or Spanish-language legal corpora, as outlined in Table~\ref{tab:all_models}. We exclude larger LLMs with more than 1 billion parameters from the fine-tuning study and instead evaluate their few-shot capabilities. We provide concise descriptions of each model and explain the rationale behind their inclusion in the study.
\begin{itemize}
    \item BERT~\citep{devlinBERTPretrainingDeep2019} is an encoder model pre-trained on a corpus of English texts. It performs masked language modeling with a vocabulary of 32K tokens and up to a sequence length of 512 tokens. Longer texts are usually truncated. We include it as an essential baseline method.
    \item Multilingual BERT (M-BERT)~\citep{devlinBERTPretrainingDeep2019} is trained in 104 languages, but otherwise, it’s the same as BERT. We include it in our study because M-BERT generates multilingual representations that are easy to fine-tune and achieve excellent performance in Spanish-language tasks~\citep{piresHowMultilingualMultilingual2019}.
    \item RoBERTa~\citep{liuRoBERTaRobustlyOptimized2019} enhances BERT with an alternative masking approach, a 10x larger corpus, and a larger vocabulary comprising 50k tokens. We include it because it generally performs better in fine-tuning tasks compared to BERT. 
    \item We include XLM-RoBERTa~\citep{conneau-etal-2020-unsupervised}, a multilingual variant of RoBERTa, as the largest model that we fine-tune. It has 270M or 560M parameters depending on the variant and is pre-trained on a corpus containing 100 languages, utilizing cross-lingual learning.
    \item DeBERTaV3~\citep{he2023debertav} enhances the original RoBERTa by separating the positional encoding of tokens from their actual content. This separation allows for better modeling of varying contexts and diverse token positions within the text, potentially leading to improved performance in downstream tasks.    
    \item Longformer~\citep{beltagyLongformerLongDocumentTransformer2020} uses a sparse attention matrix (windowed and dilated) to increase the input sequence length from 512 to 4096 tokens without quadratic growth of computational complexity. This makes it an interesting model for classifying longer clauses in the proposed corpus.
    \item Legal-BERT~\citep{chalkidis-etal-2020-legal} is pre-trained on legal corpora that include legislation, court cases, and contracts in English. However, it is otherwise similar to the original BERT. We include it because its legal vocabulary may be more adequate, even though it is trained on a different legal system. 
    \item RoBERTalex~\citep{gutierrez-fandinoSpanishLegaleseLanguage2021} is a Spanish-language model that was fine-tuned on 8.9GB of Spanish legal documents. However, it has yet to be evaluated on appropriate legal tasks, such as the Chilean Terms of Services.
    \item BETO~\citep{caneteSpanishPretrainedBERT2023}, a BERT-based model, is pre-trained on a vast corpus of Spanish texts. We include it as it is designed to handle tasks in Spanish, similar to BERT’s role in English, and occasionally even surpasses multilingual models on certain Spanish-language tasks.
    \item We include DistilBETO~\citep{canete-etal-2022-albeto}, a knowledge distilled version of BETO. This smaller and faster model is expected to retain approximately 90\% of its performance.
\end{itemize}

\begin{table*}[b]
\centering
\resizebox{\textwidth}{!}{
    \begin{tabular}{|l|c|c|c|c|c|}
        \hline
        \multicolumn{6}{|c|}{\textbf{Models selected for fine-tuning.}} \\
        \hline
        \textbf{Name} & \textbf{Params.} & \textbf{Vocab.} & \textbf{Seq. Len.} & \textbf{Train Data} & \textbf{Languages} \\
        \hline
        BERT~\citep{devlinBERTPretrainingDeep2019} & 110M & 32K & 512 & 16GB & English \\
        M-BERT~\citep{devlinBERTPretrainingDeep2019} & 110M & 32K & 512 & 16GB & 104 languages \\
        Legal-BERT~\citep{chalkidis-etal-2020-legal} & 110M & 32K & 512 & 12GB & English (legal) \\
        RoBERTa~\citep{liuRoBERTaRobustlyOptimized2019} & 125M & 50K & 512 & 160GB & English \\
        XLM-RoBERTa~\citep{conneau-etal-2020-unsupervised} & 278M/560M & 250K & 512 & 2.5TB & 100 languages \\
        RoBERTalex~\citep{gutierrez-fandinoSpanishLegaleseLanguage2021} & 278M & 50k & 512 & 8.9GB & Spanish (legal) \\
        DeBERTaV3~\citep{he2023debertav} & 86M & 128K & 512 & 160GB & English \\
        Longformer~\citep{beltagyLongformerLongDocumentTransformer2020} & 149M & 50K & 4096 & 160GB & English \\
        BETO~\citep{caneteSpanishPretrainedBERT2023} & 110M & 32K & 512 & 17.2GB & Spanish \\
        DistilBETO~\citep{canete-etal-2022-albeto} & 66M & 32K & 512 & 17.2GB & Spanish \\
        \hline
        \hline
        \multicolumn{6}{|c|}{\textbf{Models selected for few-shot learning.}} \\
        \hline
        \textbf{Name} & \textbf{Params.} & \textbf{Vocab.} & \textbf{Context} & \textbf{Train Tokens} & \textbf{Languages} \\
        \hline
        GPT-4o-2024-08-06~\citep{openaiGPT4TechnicalReport2024} & unknown & 200k & 128k & unknown & multilingual \\
        GPT-4o-mini-2024-07-18~\citep{openaiGPT4TechnicalReport2024} & unknown & 200k & 128k & unknown & multilingual \\
        Qwen2.5~\citep{yangQwen2TechnicalReport2024} & 72b, 7b & 152k & 128k & 18T & 30 languages \\
        LLama3.1~\citep{dubeyLlamaHerdModels2024} & 70b, 8b & 128k & 128k & 15T & 8 languages \\
        Mistral-Small~\citep{mistralaiteamAIAbundance2024} & 22b & 32k & 32k & unknown & 10 languages \\
        Mistral-Nemo~\citep{mistralaiteamMistralNeMo2024} & 12b & 128k & 128k & unknown & 9 languages \\
        Gemma2~\citep{gemmateamGemma2Improving2024} & 27b, 9b & 256k & 8k & 13T, 8T & primarily English \\
        \hline
    \end{tabular}
}
\caption{Overview of Large Language Models selected for fine-tuning or few-shot learning.}
\label{tab:all_models}
\end{table*}

For the detection tasks, we use a binary cross-entropy loss. For the multi-label classification task with $n$ classes, we use the binary cross entropy loss with logits, where a binary vector of length $n$ encodes the possible classes.
We report the micro-F1 and macro-F1 scores for all models and datasets since the classes are highly imbalanced. Besides these, we report precision and recall metrics for detailed inspection. To calculate the multi-label metrics, we convert the networks' sigmoid output signal to binary predictions for the $n$ classes~\citep{chalkidis-etal-2022-lexglue}.

\subsection{Prompting Large Language Models}\label{sec:prompt}

The evaluation of the few-shot capabilities of Large Language Models focuses on models that explicitly support multilingual data and have sufficiently long context lengths to process the constructed long few-shot queries. This excludes models that can only process fewer than 8,000 tokens at a time. We categorize the selected LLMs into commercial or open models. For an overview, refer to Table~\ref{tab:all_models}.

\textbf{Commercial models.} GPT-4 Omni and GPT-4 Omni Mini~\citep{openaiGPT4oSystemCard2024} are multilingual LLMs by OpenAI. We select these over the competitors due to their high performance~\citep{10.5555/3692070.3692401}, especially on multilingual benchmarks such as Multilingual Grade School Math (MGSM)~\citep{shi2023language} and a translated variant of Massive Multitask Language Understanding (MMLU)~\citep{hendrycks2021measuring}. We evaluate the model snapshots "GPT-4o-2024-08-06" and "GPT-4o-mini-2024-07-18".

\textbf{Open models.} The availability of open models changes rapidly, making it challenging to conduct a comprehensive evaluation. We include models published before November 2024. Our experiments focus on multilingual models and their variants with between 7 billion and 72 billion parameters to cover a wide range of model sizes. Smaller models may be useful for quickly screening large quantities of data, while larger models can then perform subsequent classification of potentially abusive clauses. Hence, we select the following open models:

\begin{itemize}
    \item Qwen2.5~\citep{yangQwen2TechnicalReport2024}, the latest iteration of Alibaba Group’s Qwen series, was chosen for its emphasis on multilingual support for approximately 30 languages and its handling of long context lengths (128k).
    \item Llama3.1~\citep{dubeyLlamaHerdModels2024} is an LLM released by Meta that provides official support for eight languages (including Spanish), enhancing its capabilities compared to Llama3. In two benchmarks, MGSM and multilingual MMLU, it occasionally outperforms GPT-3.5 and even surpasses GPT-4~\citep{dubeyLlamaHerdModels2024}.
    \item Gemma2~\citep{gemmateamGemma2Improving2024} is an open model by Google and was released with up to 27b parameters. Although it does not explicitly support multiple languages, its high performance in multilingual tasks~\citep{abdin2024phi} warrants its inclusion. 
    \item Mistral Small~\citep{mistralaiteamAIAbundance2024} and Nemo~\citep{mistralaiteamMistralNeMo2024} are smaller and more efficient multilingual models that could potentially compete with other smaller models like Gemma2 9b or Llama3.1 8b.
\end{itemize}

We first based our prompt on a template that was initially proposed by LegalBench~\citep{10.5555/3666122.3668037}. 
However, preliminary experimentation showed limited generalization with the different models. Errors included hallucinated class labels, extremely long answers, and high variability in prompt length in higher few-shot counts, leading to slow responses.
Hence, we adapted the prompt in two ways to mitigate some of the problems: i) we prompt more concise outputs, ii) and we sample more representative few-shot instances. 
More advanced prompting techniques, such as automated reasoning or the chain-of-thought method~\citep{shi2023language} or the incorporation of structured knowledge via retrieval-augmented generation~\citep{cuiChatlawMultiAgentCollaborative2024}, as well as the qualitative evaluation of model reasoning, are considered as interesting future work.

We describe our prompt in the following. A header describes the task and provides a list with possible labels to choose from:

\begin{quote}

   Given the following examples of clauses and their labels, predict the last clause by their label. Respond with the correct labels only. Do not explain your decision. Here are all possible options, followed by typical examples.\footnote{We use the Spanish translation: "Dados los siguientes ejemplos de cláusulas chilenas y sus 
    etiquetas, prediga la última cláusula según su etiqueta. 
    Responda sólo con las etiquetas correctas. No explique su
    decisión. A continuación, se indican todas las opciones 
    posibles, seguidas de ejemplos típicos."}
    
\end{quote}

This is then followed by one of the two following sentences, depending on whether the task is classification or detection.

\begin{quote}
    A clause can have between one and three labels.\\
    A clause can have only one label.\footnote{We use the Spanish translations: "Una cláusula puede tener entre una y tres etiquetas." or "Una cláusula sólo puede tener una etiqueta."}
\end{quote}

This is followed by a list of possible clauses to predict. For instance, when classifying potentially illegal clauses, we add the following list:

\begin{quote}

    ['na', 'lpc pro', 'cc rc', 'lpc', 'lpc int', 'lpc jus']
    
\end{quote}

This header is then followed by $n \in [1,5,10]$ examples of clauses with their corresponding labels, depending on the task and the available data in our corpus. Each class is represented at least $n$ times. However, since clauses are randomly sampled from multi-label data, there may be more examples by chance. 

The clauses in the Chilean ToS dataset have varying lengths. Some few-shot examples of extremely long clauses may pose a challenge to LLMs. Hence, we implement representative sampling that reduces the sampling noise when constructing the prompt and avoids outliers. First, we perform a binning of the clauses into $n$ bins based on their sequence lengths. We then sample representative lengths by selecting each bin's median-length sample. This sampling strategy ensures an equal distribution of sequence lengths, especially in higher shot counts, and avoids over-sampling from the long tail of the distributions shown in Fig.~\ref{fig:token_stats}. 
It can be observed that an increasing number of examples $n$ per class may significantly increase the predictive performance of some models, see Section~\ref{sec:results}. 

After the list of examples, we inject the clause to classify at the position \{\{\}\} and prompt the model to generate the correct label:

\begin{quote}
    
    Clause: \{\{\}\}\\
    Label:\footnote{We use the Spanish translations "Cláusula" and "Etiqueta"}
    
\end{quote}

\section{Experiments}\label{sec:results}

This section describes the experimental setup and presents the results for the task detection in Sec.~\ref{sec:detect}, and for classification in Sec.~\ref{sec:classify}. A detailed error analysis follows in Sec.~\ref{sec:errors}.

\textbf{Experimental setup.} We obtain pre-trained models from either the Huggingface Hub\footnote{\url{https://huggingface.co/models}, transformers version 4.44.0} or the Ollama Library\footnote{\url{https://ollama.com/library}, Ollama version 0.3.11}. For few-shot learning, we select "instruct" variants that are tuned to follow prompts. We use 8-bit quantization, because this level of quantization increases throughput considerably, while still recommended for processing long texts because its negative effect on performance is negligible~\citep{liEvaluatingQuantizedLarge2024}. We construct the few-shot prompts from the combined training and validation datasets. For each model, we perform runs with three different random seeds and report their average test set scores with standard deviation.
We fine-tune the smaller models using a binary cross entropy loss with multi-label logits and use the Adam optimizer~\citep{kingmaADAMMETHODSTOCHASTIC2015} with a learning rate of 3e-5 and a batch size of 32 with up to 50 epochs. Where necessary, we apply gradient accumulation to obtain the target batch size. We regularize the training further with an early-stopping mechanism set to 10 epochs of patience. It tracks the macro-F1 score of the validation split because the datasets are imbalanced. Our fine-tuning experiments extend the code base published by LexGlue~\citep{chalkidis-etal-2022-lexglue}\footnote{\url{https://github.com/coastalcph/lex-glue}, commit 20ad76a} and accelerate training using mixed precision on NVIDIA Ada Lovelace GPUs. 

Since the datasets are highly imbalanced we do not only report the micro-F1 score ($\mu$-F1), which provides a global measure but may be dominated by the majority classes. We also report the macro-F1 (m-F1) score, which allows us to evaluate models on the classes of each dataset equally, highlighting the performance of minority classes. For the best-performing fine-tuned models we additionally break down the results into precision and recall for each classification task, see Sec.~\ref{sec:errors}. Our study aggregates the results from a total of 426 experiments. 

\subsection{Detection}\label{sec:detect}

We demonstrate the ability of LLMs to detect potentially abusive clauses in the Chilean ToS corpus.
To this end, we conducted large-scale experiments with a diverse set of models on the three detection datasets of the corpus and reported the $\mu$-F1 and m-F1 scores and their standard deviation per model over three runs.
The results in Tab.~\ref{tab:res_fs_detect} are subdivided into four groups of models: i) an SVM baseline that was trained from scratch, ii) six transformer models pre-trained on English, iii) eight transformer models pre-trained with Spanish or multiple languages, and iv) four LLMs queried via few-shot learning. 

\begin{table*}[t]
    \centering
    \resizebox{\textwidth}{!}{
        \begin{tabular}{|l|cc|cc|cc|cc|cc|cc|cc|c|c}
            \hline
             \multirow{2}{*}{\bf Method}  & \multicolumn{2}{c|}{\textbf{Illegal Detection}} & \multicolumn{2}{c|}{\textbf{Dark Detection}} & \multicolumn{2}{c|}{\textbf{Gray Detection}} \\
              & $\mu$-F1 & m-F1 & $\mu$-F1 & m-F1 & $\mu$-F1 & m-F1 \\ 
\hline
SVM+TF-IDF 		& 0.94 $\pm$0.00 & 0.85 $\pm$0.00 & \textbf{0.96} $\pm$0.00 & 0.79 $\pm$0.04 & 0.94 $\pm$0.00 & 0.70 $\pm$0.03 \\
\hline
    BERT cased & 0.95 $\pm$0.01 & 0.87 $\pm$0.02 & \underline{0.95} $\pm$0.00 & 0.76 $\pm$0.02 & \textbf{0.96} $\pm$0.00 & \underline{0.77} $\pm$0.03 \\
    BERT uncased & \textbf{0.96} $\pm$0.00 & \textbf{0.89} $\pm$0.00 & \underline{0.95} $\pm$0.00 & \underline{0.77} $\pm$0.02  & 0.95 $\pm$0.00 & 0.75 $\pm$0.01 \\
    RoBERTa & 0.94 $\pm$0.01 & 0.85 $\pm$0.02 & 0.94 $\pm$0.01 & 0.58 $\pm$0.15 & 0.93 $\pm$0.00 & 0.51 $\pm$0.05 \\
    Legal-BERT & 0.95 $\pm$0.00 & 0.87 $\pm$0.00 & 0.94 $\pm$0.01 & 0.68 $\pm$0.17 & 0.95 $\pm$0.00 & 0.75 $\pm$0.02 \\
    DeBERTaV3 & 0.95 $\pm$0.00 & 0.88 $\pm$0.00 & 0.94 $\pm$0.01 & 0.58 $\pm$0.18 & 0.94 $\pm$0.01 & 0.55 $\pm$0.11 \\
    Longformer & 0.95 $\pm$0.00 & 0.87 $\pm$0.01 & 0.93 $\pm$0.00 & 0.50 $\pm$0.03 & 0.93 $\pm$0.00 & 0.50 $\pm$0.03 \\
\hline
    M-BERT cased & \underline{0.95} $\pm$0.01 & 0.87 $\pm$0.01 & \underline{0.95} $\pm$0.00 & 0.76 $\pm$0.01 & 0.95 $\pm$0.00 & 0.76 $\pm$0.02 \\
    M-BERT uncased & \underline{0.95} $\pm$0.00 & \underline{0.88} $\pm$0.00 & \underline{0.95} $\pm$0.00 & 0.78 $\pm$0.03 & 0.95 $\pm$0.01 & 0.77 $\pm$0.04 \\
    XLM-RoBERTa-base & \underline{0.95} $\pm$0.00 & 0.86 $\pm$0.01 & 0.94 $\pm$0.01 & 0.65 $\pm$0.12 & 0.94 $\pm$0.01 & 0.67 $\pm$0.16 \\
    XLM-RoBERTa-large & 0.87 $\pm$0.00 & 0.47 $\pm$0.00 & 0.93 $\pm$0.00 & 0.48 $\pm$0.00& 0.93 $\pm$0.00 & 0.48 $\pm$0.00 \\
    RoBERTalex & 0.94 $\pm$0.01 & 0.86 $\pm$0.01 & \underline{0.95} $\pm$0.00 & 0.76 $\pm$0.01 & 0.95 $\pm$0.00 & 0.71 $\pm$0.01 \\
    BETO cased & \underline{0.95} $\pm$0.00 & \underline{0.88} $\pm$0.00 & \underline{0.95} $\pm$0.00 & 0.78 $\pm$0.01 & \textbf{0.96} $\pm$0.01 & \textbf{0.79} $\pm$0.03 \\
    BETO uncased & \underline{0.95} $\pm$0.00 & \underline{0.88} $\pm$0.01 & \underline{0.95} $\pm$0.00 & \textbf{0.80} $\pm$0.00 & 0.95 $\pm$0.00 & 0.77 $\pm$0.02 \\
    DistilBETO & \underline{0.95} $\pm$0.00 & \underline{0.88} $\pm$0.01  & \underline{0.95} $\pm$0.00 & 0.78 $\pm$0.03 & 0.95 $\pm$0.00 & 0.76 $\pm$0.02 \\
\hline
\hline
  GPT-4o-mini & 0.81 $\pm$0.00 & 0.68 $\pm$0.00  & \underline{0.86} $\pm$0.00  & \underline{0.66} $\pm$0.00  & \underline{0.78} $\pm$0.00 & \underline{0.60} $\pm$0.00 \\
  Llama3.1 70b & \underline{0.86} $\pm$0.00 & \underline{0.73} $\pm$0.00 & 0.84 $\pm$0.01 & 0.65 $\pm$0.01 & 0.07 $\pm$0.00 & 0.08 $\pm$0.00 \\
  Gemma2 27b & 0.79 $\pm$0.00 & 0.68 $\pm$0.00 & 0.71 $\pm$0.00 & 0.56 $\pm$0.00 & 0.09 $\pm$0.00 & 0.10 $\pm$0.00 \\
  Llama3.1 8b & 0.62 $\pm$0.01 & 0.54 $\pm$0.01 & 0.57 $\pm$0.00 & 0.46 $\pm$0.00 & 0.08 $\pm$0.00 & 0.09 $\pm$0.00 \\

\hline
        \end{tabular}
    }
    \caption{Experimental results for the three detection datasets. For convenience, we print the \textbf{overall best} results in bold and underline the results that are \underline{best per group}. The results show that detection is feasible. Moreover, they indicate a difficulty-ranking of the corpora with the Illegal Detection dataset on the one hand being easier to predict and Dark Detection and Gray Detection on the other hand harder, resulting in lower detection rates. }
    \label{tab:res_fs_detect}
\end{table*}

\textbf{On the Illegal Detection dataset}, the English-language transformer BERT achieve the highest F1 scores, with 96\% $\mu$-F1 and 89\% m-F1. Compared to the SVM baseline, BERT uncased obtains a 2\% higher $\mu$-F1 score and 4\% higher m-F1 score. 
The Spanish and multilingual transformers, such as the BETO models, nearly match BERT. However, XLM-Roberta falls behind, and its large variant fails to learn the task. Surprisingly, RoBERTalex tails by 3\% despite Spanish-language legalese pre-training.
Few-shot learning is not effective and the best models are behind by 10\% in $\mu$-F1 and 16\% $m$-F1.
    
\textbf{On the Dark Detection dataset}, the baseline SVM delivers the highest $\mu$-F1 Score of 96\% and  BETO uncased the highest m-F1 Score of 80\%. The Dark Clauses are harder to detect than Illegal Clauses. This may be due to their relatively lower occurrence in the corpus and their imbalance, leading to lower m-F1 scores.
Most English transformers exhibit a very high variance in m-F1 scores and significant drops in performance. The regularization may not completely eliminate overfitting to more frequent classes, possibly leading to this problematic higher variance.
However, more appropriate pre-training remedies these problems as multilingual or Spanish-language pre-training demonstrates. RoBERTalex performs on par with M-BERT cased and only falls behind BETO and M-BERT uncased by 2\% m-F1 Score.
Finally, few-shot learning is ineffective and the best models are behind by 10\% in $\mu$-F1 and 14\% $m$-F1.
    
\textbf{On the Gray Detection dataset}, we observe lower $m$-F1 scores due to the complexity of this ambiguous dataset, as seen by the lower results of the comparably simplistic SVM approach. The Spanish-language BETO cased achieves an $\mu$-F1 Score of 96\% and an m-F1 Score of 79\%, significantly outperforming both the Spanish legalese RoBERTalex model as well as the multilingual XLM-RoBERTa and M-BERT variants. The advantage of pre-training is confirmed by the weak performance of most English-language models where only the (Legal-) BERT variants remain competitive.
Few-shot learning remains ineffective and the best models are behind by 18\% in $\mu$-F1 and 19\% $m$-F1.

\textbf{In summary}, our proposed detection datasets present their unique challenges and our analysis emphasizes the importance of selecting a suitable model architecture and dataset characteristics, such as pre-training corpora. However, the strong performance of the SVM and small, appropriately pre-trained transformers demonstrates, that the detection of potentially abusive clauses in the Chilean ToS corpus is possible, with the rates of $96\%$ $\mu$-F1 and $89\%$ m-F1 on Illegal Detection, $96\%$ $\mu$-F1 and $80\%$ m-F1 on Dark Detection, $96\%$ $\mu$-F1 and $79\%$ m-F1 on Gray Detection. This allows us to separate the detection from subsequent classification, which we will discuss in the next section. 

\subsection{Classification}\label{sec:classify}
In our proposed approach for the corpus, classification follows on detection. This chapter demonstrates the capabilities of a diverse set of models to classify potentially abusive clauses for each of the three subsets Illegal Classification, Dark Classification, and Gray Classification, utilizing fine-tuning or few-shot learning.

We extend our experiments with additional LLMs for a more representative selection of available models (see Tab.~\ref{tab:all_models}) and provide results for few-shot prompting with $n\in[1,5,10]$ for both GPT models. We perform each experiment three times and report the $\mu$- and m-F1 scores with standard deviation in Tab.~\ref{tab:class_combined_results}. 

\begin{table*}[t]
    \centering
    \resizebox{\textwidth}{!}{
        \begin{tabular}{|l|cc|cc|cc|cc|cc|cc|cc|c|c}
            \hline
             \multirow{2}{*}{\bf Method}  & \multicolumn{2}{c|}{Illegal Classification} & \multicolumn{2}{c|}{Dark Classification} & \multicolumn{2}{c|}{Gray Classification} \\
             & $\mu$-F1 & m-F1 & $\mu$-F1 & m-F1 & $\mu$-F1 & m-F1 \\
            \hline
\hline
SVM+TF-IDF 		& 0.63 $\pm$0.01 & 0.47 $\pm$0.01 & 0.69 $\pm$0.01 & 0.48 $\pm$0.04 & 0.51 $\pm$0.00 & 0.55 $\pm$0.00 \\
\hline
\hline
BERT uncased & 0.74 $\pm$0.01 & 0.62 $\pm$0.02 & 0.73 $\pm$0.04 & 0.45 $\pm$0.05 & 0.42 $\pm$0.00 & 0.39 $\pm$0.04 \\
BERT cased & 0.70 $\pm$0.03 & 0.57 $\pm$0.02 & 0.71 $\pm$0.03 & 0.41 $\pm$0.03 & 0.43 $\pm$0.06 & 0.44 $\pm$0.08 \\
RoBERTa & 0.70 $\pm$0.05 & 0.56 $\pm$0.08 & 0.72 $\pm$0.06 & 0.49 $\pm$0.09 & 0.47 $\pm$0.02 & 0.50 $\pm$0.01 \\
Legal-BERT & \underline{0.75} $\pm$0.01 & 0.62 $\pm$0.01 & 0.74 $\pm$0.02 & 0.42 $\pm$0.01 & 0.42 $\pm$0.10 & 0.38 $\pm$0.13 \\
DeBERTaV3 & 0.73 $\pm$0.02 & \underline{0.66} $\pm$0.03 & \underline{0.76} $\pm$0.03 & \underline{0.50} $\pm$0.09 & \underline{0.53} $\pm$0.02 & \underline{0.57} $\pm$0.03 \\
Longformer & 0.70 $\pm$0.02 & 0.54 $\pm$0.04 & 0.64 $\pm$0.07 & 0.41 $\pm$0.08 & 0.47 $\pm$0.07 & 0.51 $\pm$0.07 \\
\hline
M-BERT cased & 0.74 $\pm$0.04 & 0.62 $\pm$0.09 & 0.77 $\pm$0.03 & 0.46 $\pm$0.02 & 0.48 $\pm$0.05 & 0.51 $\pm$0.06 \\
M-BERT uncased & 0.74 $\pm$0.01 & 0.64 $\pm$0.02 & 0.75 $\pm$0.02 & 0.43 $\pm$0.00 & \underline{0.57} $\pm$0.03 & 0.57 $\pm$0.04 \\
XLM-RoBERTa-base & 0.75 $\pm$0.03 & 0.64 $\pm$0.06 & 0.76 $\pm$0.07 & \underline{0.55} $\pm$0.16 & 0.53 $\pm$0.04 & 0.56 $\pm$0.03 \\
XLM-RoBERTa-large & \textbf{0.78} $\pm$0.01 & \textbf{0.70} $\pm$0.01 & 0.78 $\pm$0.02 & 0.52 $\pm$0.06 & 0.52 $\pm$0.17 & 0.47 $\pm$0.31 \\
RoBERTalex & 0.74 $\pm$0.02 & 0.63 $\pm$0.03 & 0.78 $\pm$0.01 & 0.45 $\pm$0.01 & 0.56 $\pm$0.02 & \underline{0.58} $\pm$0.02 \\
BETO cased & 0.74 $\pm$0.01 & 0.63 $\pm$0.04 & \textbf{0.80} $\pm$0.01 & 0.50 $\pm$0.06 & 0.55 $\pm$0.03 & 0.56 $\pm$0.04 \\
BETO uncased & 0.75 $\pm$0.02 & 0.66 $\pm$0.05 & 0.78 $\pm$0.03 & 0.47 $\pm$0.02 & 0.51 $\pm$0.03 & 0.53 $\pm$0.03 \\
DistilBETO & 0.73 $\pm$0.01 & 0.61 $\pm$0.02 & 0.77 $\pm$0.02 & 0.45 $\pm$0.02 & 0.42 $\pm$0.03 & 0.44 $\pm$0.03 \\
\hline
\hline
GPT-4o  (1-shot) & 0.46 $\pm$0.01 & 0.46 $\pm$0.01 & 0.69 $\pm$0.01 & 0.52 $\pm$0.01 & 0.56 $\pm$0.01 & 0.59 $\pm$0.00 \\
GPT-4o (5-shot) & 0.61 $\pm$0.01 & \underline{0.68} $\pm$0.00 & \underline{0.74} $\pm$0.01 & 0.59 $\pm$0.00 & 0.61 $\pm$0.01 & 0.61 $\pm$0.01 \\
GPT-4o (10-shot) & 0.62 $\pm$0.00 & \underline{0.68} $\pm$0.01 & - & - & \textbf{0.63} $\pm$0.01 & \textbf{0.66} $\pm$0.00 \\

GPT-4o-mini (1-shot) & 0.51 $\pm$0.01 & 0.49 $\pm$0.01 & \underline{0.74} $\pm$0.01 & \textbf{0.60} $\pm$0.04 & 0.57 $\pm$0.01 & 0.58 $\pm$0.01 \\
GPT-4o-mini (5-shot)  & 0.61 $\pm$0.01 & 0.61 $\pm$0.01 & 0.73 $\pm$0.02 & \textbf{0.60} $\pm$0.03 & 0.61 $\pm$0.00 & \textbf{0.66} $\pm$0.00 \\
GPT-4o-mini (10-shot) & \underline{0.64} $\pm$0.01 & 0.65 $\pm$0.01 & - & - & 0.61 $\pm$0.01 & 0.65 $\pm$0.01 \\
\hline
  Qwen2.5 72b & 0.52 $\pm$0.02 & \underline{0.50} $\pm$0.02 & \underline{0.70} $\pm$0.01 & 0.53 $\pm$0.01 & 0.56 $\pm$0.02 & 0.58 $\pm$0.02 \\
  Qwen2.5 7b & 0.51 $\pm$0.01 & 0.41 $\pm$0.02 & 0.67 $\pm$0.01 & \underline{0.59} $\pm$0.01 & 0.52 $\pm$0.01 & 0.51 $\pm$0.02 \\
  Llama3.1 70b & 0.43 $\pm$0.03 & 0.44 $\pm$0.04 & 0.64 $\pm$0.01 & 0.54 $\pm$0.01 & \underline{0.58} $\pm$0.02 & \underline{0.61} $\pm$0.01 \\
  Llama3.1 8b & 0.24 $\pm$0.01 & 0.21 $\pm$0.01 & 0.55 $\pm$0.04 & 0.46 $\pm$0.04 & 0.42 $\pm$0.02 & 0.44 $\pm$0.02 \\
  Mistral-Small 22b & 0.45 $\pm$0.03 & 0.39 $\pm$0.02 & 0.64 $\pm$0.02 & 0.48 $\pm$0.04 & 0.20 $\pm$0.03 & 0.21 $\pm$0.05 \\
  Mistral-Nemo 12b & 0.36 $\pm$0.05 & 0.34 $\pm$0.04 & 0.62 $\pm$0.02 & 0.50 $\pm$0.04 & 0.45 $\pm$0.03 & 0.49 $\pm$0.01 \\
  Gemma2 27b & \underline{0.57} $\pm$0.04 & 0.48 $\pm$0.03 & \underline{0.70} $\pm$0.01 & 0.55 $\pm$0.01 & 0.52 $\pm$0.01 & 0.57 $\pm$0.01 \\
  Gemma2 9b & 0.44 $\pm$0.02 & 0.36 $\pm$0.02 & 0.67 $\pm$0.03 & 0.53 $\pm$0.03 & 0.51 $\pm$0.01 & 0.53 $\pm$0.02 \\

            \hline
        \end{tabular}
    }
    \caption{Experimental results for the classification tasks. We group fine-tuned LLMs into models, that were pre-trained in English, or those, that were pre-trained in Spanish or more languages. We group LLMs for few-shot learning into commercial or open models. Few-shot evaluations are performed with $n=1$. In addition, we include results for $n=5$ and $n=10$ for GPT-4o and -mini. The other LLMs did not benefit from higher $n$. The SVM with TF-IDF features serves as the baseline. For convenience, we print the \textbf{overall best} results in bold and underline the results that are \underline{best per group}.}
    \label{tab:class_combined_results}
\end{table*}

\textbf{The Illegal Classification dataset} makes up about 45\% of the corpus and consists of classes with many examples each. In our experiments, we observe that the largest fine-tuned model XLM-RoBERTa-large shows the highest $\mu$-F1 score of $78\%$ and m-F1 score of $70\%$, and significantly outperforms few-shot learners. This may be due to the circumstance, that Illegal Clauses form a large enough training dataset to generalize well to unseen examples, and that Illegal Clauses infringe specific laws using specific wording, such as references to non-Chilean jurisdictions, that aids the classifier model.
From the remaining fine-tuned models, within the group of English-language models, DeBERTaV3 leads significantly in the m-F1 score by 4\%, demonstrating the benefits of its architectural advances compared to the baseline transformers like BERT. However, the broad Spanish-language pre-training of BETO uncased, and to a lesser extent of RoBERTalex, compensates for DeBERTaV3's architectural advantages. Overall, the fine-tuned Spanish-language or multilingual models perform on par or better than English-language models. 

Few-shot learners achieve competitive $m$-F1 scores of $68\%$ for GPT-4o (5- or 10-shots), which is only surpassed by XLM-RoBERTa-large. On $\mu$-F1, however, the few-shot learners fall behind by up to $14\%$.
The open LLMs show mixed results with Gemma2 27b and Gwen2.5 72b performing strong in the 1-shot scenario and smaller models and both Llama3.1 variants fall behind. We observe that only the GPT-4o variants can leverage higher $n$ of 5 and 10, enabling them to compete on $m$-F1 score with fine-tuned models.

\textbf{The Dark Classification clauses} are manifestly abusive clauses defined by Article 16 LPC. Here, difficulties may arise from clause reformulation and other intra-class variability. In addition, the class imbalance combined with a multi-label classification may prove difficult.
We observe different types of approaches leading in $m$-F1 or $\mu$-F1 scores.
In $\mu$-F1, the highly specialized Spanish-language models BETO and RoBERTalex lead by up to 4\% over the best English-language model DeBERTaV3 and 6\% over the best few-shot learners GPT-4o/-mini. However, while they dominate on the $\mu$-F1 score, the GPT-variants and several open LLMs perform stronger in the $m$-F1 metric. For GPT-4o-mini, this lead is $5$ to $10\%$ $m$-F1 Score compared to the best fine-tuned models. The open LLMs Gemma2 27B and both Qwen2.5 variants are competitive in $m$-F1 scores but disappoint in the $\mu$-F1 metric. These interesting differences between learning approaches may be due to fewer available training data for fine-tuning, while the GPT models can utilize their world knowledge to predict rare classes from only a few available examples.

\textbf{The Gray Classification dataset} is ambiguous and hard to analyze for ML models and expert annotators alike. Their identification may depend on contextual facts not present in the clauses themselves. However, especially this knowledge of contextual facts may be why the larger few-shot learners surpass the smaller fine-tuned LLMs in both $\mu$- and $m$-F1. For example, we see GPT-4o (10-shots) and GPT-4o-mini (5- and 10-shots) significantly outperforming any fine-tuned model by between $6\%$ ($\mu$-F1) and $9\%$ (m-F1). The open model Llama3.1 70b also achieves higher scores of $58\%$ $\mu$-F1 and $61\%$ m-F1. 
An increase in the number of examples (shots) provided in the prompts helps capture the dataset’s diversity, enabling the GPTs to make accurate predictions of subtle variations. These results demonstrate that few-shot learners can generalize well on the low-data, highly ambiguous, multi-label dataset. While our experiments with larger shot counts for open models were not successful, the positive results from GPT-4o models indicate great potential with a more specialized prompting technique. We briefly discuss cost and privacy considerations in Appendix~\ref{secA2}.

In contrast, we observe that fine-tuning on the smaller training dataset of Gray Classification is not sufficient for good generalization. Still, RoBERTalex benefits from its domain-specific pre-training and delivers $\mu$-F1 Score of $58\%$, and M-BERT presents a balanced performance of $57\%$ for $\mu$-F1 and $m$-F1 alike. Surprisingly, the SVM with TF-IDF features performs better than most English-language models.

\textbf{In summary}, our study shows that ML can significantly aid in the classification of potentially abusive clauses on the Chilean ToS corpus. The differences in performance between the Illegal, Dark, and Gray Classification indicate the importance of a suitable model selection and knowledge of contextual facts, given the datasets' high levels of ambiguity (see Fig.~\ref{fig:co-occurence}) and diverse training data sizes (see Tab.~\ref{tab:classify_stats}). Leveraging language- and domain-specific pre-training, and modern architectures and training methods, as well as in-context learning with large shot counts, can significantly improve classification.

\subsection{Error Analysis}\label{sec:errors}

This section conducts a more profound performance analysis to show the difficulties in the proposed datasets.
To this end, we present class-wise precision and recall results. 
Figure~\ref{tab:clsrep:combined} shows classification reports for the three datasets. We analyze the fine-tuned models with the highest m-F1 score per dataset: BETO uncased for Illegal Classification, XLM-RoBERTa-large for Dark Classification, and RoBERTalex for Gray Classification.

\begin{table}
\caption{Classification Reports for the best performing fine-tuned models on the datasets Illegal Classification, Dark Classification, and Gray Classification.}
\label{tab:clsrep:combined}
\begin{tabular}{l}
\toprule
\textbf{Illegal Classification, BETO uncased.} \\
\midrule
\begin{tabular}{llllr}
 & Precision & Recall & F1-score & Support \\
\midrule
na & 0.81 $\pm$ 0.02 & 0.81 $\pm$ 0.01 & 0.81 $\pm$ 0.01 & 86 \\
lpc pro & 0.68 $\pm$ 0.03 & 0.79 $\pm$ 0.04 & 0.73 $\pm$ 0.02 & 41 \\
rc & 0.89 $\pm$ 0.03 & 0.82 $\pm$ 0.00 & 0.86 $\pm$ 0.02 & 17 \\
lpc & 0.65 $\pm$ 0.05 & 0.51 $\pm$ 0.04 & 0.57 $\pm$ 0.04 & 13 \\
lpc int & 1.00 $\pm$ 0.00 & 0.67 $\pm$ 0.14 & 0.79 $\pm$ 0.10 & 12 \\
lpc jus & 0.33 $\pm$ 0.29 & 0.14 $\pm$ 0.13 & 0.19 $\pm$ 0.17 & 12 \\
avg / total & 0.73 $\pm$ 0.07 & 0.62 $\pm$ 0.06 & 0.66 $\pm$ 0.06 & 181 \\
\end{tabular} \\
\midrule
\textbf{Dark Classification, XLM-RoBERTa-large.} \\
\midrule
\begin{tabular}{llllr}
 & Precision & Recall & F1-score & Support \\
\midrule
ltd & 0.85 $\pm$ 0.02 & 0.86 $\pm$ 0.05 & 0.85 $\pm$ 0.03 & 47 \\
cr & 0.80 $\pm$ 0.01 & 0.91 $\pm$ 0.02 & 0.85 $\pm$ 0.01 & 29 \\
nod & 0.84 $\pm$ 0.05 & 0.69 $\pm$ 0.05 & 0.76 $\pm$ 0.03 & 12 \\
er & 0.76 $\pm$ 0.21 & 0.21 $\pm$ 0.05 & 0.32 $\pm$ 0.05 & 11 \\
ch & 0.50 $\pm$ 0.50 & 0.33 $\pm$ 0.33 & 0.36 $\pm$ 0.31 & 3 \\
ter & 0.00 $\pm$ 0.00 & 0.00 $\pm$ 0.00 & 0.00 $\pm$ 0.00 & 3 \\
avg / total & 0.62 $\pm$ 0.13 & 0.50 $\pm$ 0.08 & 0.52 $\pm$ 0.07 & 105 \\
\end{tabular} \\
\midrule
\textbf{Gray Classification, RoBERTalex.} \\
\midrule
\begin{tabular}{llllr}
 & Precision & Recall & F1-score & Support \\
\midrule
des risk & 0.64 $\pm$ 0.05 & 0.45 $\pm$ 0.09 & 0.53 $\pm$ 0.07 & 25 \\
bfe & 0.57 $\pm$ 0.04 & 0.38 $\pm$ 0.03 & 0.45 $\pm$ 0.01 & 23 \\
des reser & 0.69 $\pm$ 0.06 & 0.73 $\pm$ 0.09 & 0.70 $\pm$ 0.03 & 17 \\
des uni & 0.55 $\pm$ 0.08 & 0.36 $\pm$ 0.12 & 0.42 $\pm$ 0.09 & 14 \\
des det & 1.00 $\pm$ 0.00 & 0.12 $\pm$ 0.05 & 0.21 $\pm$ 0.08 & 11 \\
des def & 1.00 $\pm$ 0.00 & 0.60 $\pm$ 0.10 & 0.75 $\pm$ 0.08 & 10 \\
des inf & 1.00 $\pm$ 0.00 & 0.50 $\pm$ 0.00 & 0.67 $\pm$ 0.00 & 8 \\
des lic & 1.00 $\pm$ 0.00 & 0.80 $\pm$ 0.00 & 0.89 $\pm$ 0.00 & 5 \\
avg / total & 0.81 $\pm$ 0.03 & 0.49 $\pm$ 0.06 & 0.58 $\pm$ 0.05 & 113 \\
\end{tabular} \\
\bottomrule
\end{tabular}
\end{table}

\textbf{On Illegal Classification}, the report indicates that especially "lpc jus" is hard to predict for BETO. This may be due to the diversity of this case of potentially abusive clauses. It includes a variety of ways that contracts limit access to justice, such as the ability to access justice directly, or infringements upon time limits, the burden of proof, presumptions, or the court competencies, e.g., the following clause restricts access to justice in Chile. However, the model predicts this instance incorrectly as the more specific "lpc pro":

\begin{quote}
    You and Nintendo agree that the state or federal courts in King County, Washington have exclusive jurisdiction over any appeal of an arbitration award and over any claim between the parties that has not been submitted to arbitration. Disputes between the parties shall be governed by this Agreement and the applicable laws of the state of Washington and the United States, without giving effect to principles of conflicts of laws that would permit the application of the law of another jurisdiction.\footnote{Translated from Spanish using DeepL.com: "Usted y Nintendo acuerdan que los tribunales estatales o federales del condado de King, Washington, tienen jurisdicción exclusiva sobre cualquier apelación de un laudo arbitral y sobre cualquier demanda entre las partes que no se haya sometido a arbitraje. Los conflictos entre las partes se regirán por este Acuerdo y las leyes aplicables del estado de Washington y de Estados Unidos, sin que se ejecuten principios de conflicto de leyes que permitan la aplicación de la ley de otra jurisdicción."}
\end{quote}

This example emphasizes the difficulty of predicting the distinction between a specific alternation of the LPC's jurisdiction rules ("lpc pro") and the more general limitation of access to justice ("lpc jus").

We record the second-worst scores for the class "lpc", which acts as a residual class and contains fewer specific clauses. In contrast, the remaining labels "na", "lpc pro", "rc" and "lpc int" are more specific and thus more reliably classified.

\textbf{On Dark Classification}, XLM-RoBERTa-large is severely challenged. The dataset is highly imbalanced. The two classes "ch" and "ter" are extremely rare, and thus, hard to learn correctly. The following example shows a clause that allows the unilateral modification of the price of a service:

\begin{quote}
    The fees we charge sellers for using our Services to sell products and services are set forth on our Selling Rates pages. We may change our selling fees from time to time. Such changes will be posted on the eBay site fourteen (14) days in advance. Prior notice is not required for temporary promotions or any changes that result in a reduction of fees.\footnote{Translated from Spanish using DeepL.com: "Las tarifas que les cobramos a los vendedores por usar nuestros Servicios para vender productos y servicios se indican en nuestras páginas de Tarifas de venta. Podemos cambiar nuestras tarifas de venta periódicamente. Dichos cambios se publicarán en el sitio de eBay con catorce (14) días de antelación. No se requiere una notificación previa para promociones temporales o cualquier cambio que tenga como consecuencia una reducción de las tarifas."}
\end{quote}

The model is not leveraging the relevant keywords for the following exemplary clause of the rare class "ch", such as "change", "fees" or "Selling Rates" in their respective Spanish original. Instead, the model predicts "cr", which refers to a unilateral change of the ToS without users being able to challenge or reject the changes.

Furthermore, the clauses are often annotated with multiple labels, e.g., as both one-sided (er) and limiting the company's liability (ltd). In consequence, the model overfits to the much more frequent "ltd" class (precision 85\%, recall 86\%) and does not recall "er" well (recall 33\%).

\textbf{On Gray Classification}, samples are not only highly ambiguous but also annotated with up to three labels per sample. The initially surprising result of perfect precision scores for four of the eight classes ("des det", "des def", "des inf" and "des lic") can be explained by their low rate of overlap with other classes. Only "des det", which also scores the lowest recall score of the eight classes, often co-occurs with other classes. The three most frequent classes "des risk", "bfe" and "des reser" co-occur at a high rate, which leads to higher confusion that expresses itself in the lower scores. An example of this issue is the following "des det" clause, which affects the processes for resolving conflicts:

\begin{quote}
In the event that the User breaches a law or the Terms and Conditions, we may warn, suspend, restrict or temporarily or permanently disable his/her account, without prejudice to other sanctions established in the particular rules of use of Mercado Libre's services.\footnote{Translated from Spanish using DeepL.com: "En caso que la Persona Usuaria incumpliera una ley o los Términos y Condiciones, podremos advertir, suspender, restringir o inhabilitar temporal o definitivamente su cuenta, sin perjuicio de otras sanciones que se establezcan en las reglas de uso particulares de los servicios de Mercado Libre."}
\end{quote}

RoBETalex incorrectly classifies this as "des reser", which allows the provider to modify the contract at their discretion. This error may be explained due to the relatively high co-occurrence of both classes in the dataset, see Fig.~\ref{fig:co-occurence} for reference.

\textbf{In summary}, our detailed analysis of precision and recall across the datasets of the Chilean ToS corpus underscores the need for addressing class imbalance and multi-label complexities in legal NLP tasks.

\section{Discussion}\label{sec:discussion}

ML is becoming an increasingly potent tool to process the large volume of information generated within the legal domain~\citep{grecoBringingOrderRealm2024}. Especially the family of transformer-based language models has come into widespread use with models ranging from millions~\citep{devlinBERTPretrainingDeep2019,conneau-etal-2020-unsupervised,chalkidis-etal-2020-legal,caneteSpanishPretrainedBERT2023,gutierrez-fandinoSpanishLegaleseLanguage2021} to billions~\citep{openaiGPT4TechnicalReport2024,yangQwen2TechnicalReport2024,dubeyLlamaHerdModels2024,gemmateamGemma2Improving2024,mistralaiteamAIAbundance2024,mistralaiteamMistralNeMo2024} of parameters. However, the main line of research has long been focused on English-language models due to the high availability of datasets~\citep{devlinBERTPretrainingDeep2019,liuRoBERTaRobustlyOptimized2019,he2023debertav,beltagyLongformerLongDocumentTransformer2020}. With the success of these methods, multilingual~\citep{conneau-etal-2020-unsupervised,devlinBERTPretrainingDeep2019} and language-specific models, e.g., the Spanish-language model BETO~\citep{caneteSpanishPretrainedBERT2023}, and domain-specific models for legal problems, e.g, the English-only Legal-BERT~\citep{chalkidis-etal-2020-legal} or recently Lawma~\citep{dominguez-olmedoLawmaPowerSpecialization2024}, gained popularity. Even though such language- and domain-specific models may perform better in their designated tasks, they are harder to develop and evaluate. For example, the Spanish legalese model RoBERTalex~\citep{gutierrez-fandinoSpanishLegaleseLanguage2021} has not yet been evaluated on suitable tasks due to a lack of high-quality domain-specific datasets. 

\cite{grecoBringingOrderRealm2024} recognize that free-access legal resources are often limited, and the lack of datasets is a challenge in legal NLP. We find this issue exacerbated by the focus on the English language and major jurisdictions, such as the European Union or the US. In the realm of consumer contract analysis, we mainly find few and mainly European law datasets~\citep{braun-matthes-2024-agb,lippiCLAUDETTEAutomatedDetector2019,ruggeriDetectingExplainingUnfairness2022,drawzeskiCorpusMultilingualAnalysis2021,dadasSupportSystemDetection2024}. Instead of new datasets, new publications often appear as curated dataset collections~\citep{chalkidis-etal-2022-lexglue,10.5555/3666122.3668037,fei-etal-2024-lawbench,niklausLEXTREMEMultiLingualMultiTask2023}. While these help drive legal NLP's method development~\citep{10.5555/3666122.3668037,akashUnfairTOSAutomated2024} their scope often remains limited. In this publication, we address the critical gap of non-English-language legal datasets that are not from the usual major jurisdictions and present a novel Spanish-language legal dataset of Chilean law. In addition, we propose a large set of strong baseline methods against which to measure future developments. This work sets the stage for future research in the legal context.

In the rest of this section, we discuss transformer-based models in Sec.~\ref{sec:rw:transformers}, relevant datasets in Sec.~\ref{sec:rw:datasets}, and conclude the discussion with an outlook in Sec.~\ref{sec:rw:outlook}.

\subsection{Transformer-based Models} \label{sec:rw:transformers}

With the introduction of the Attention mechanism~\citep{10.5555/3295222.3295349}, a new wave of model architectures emerged, with BERT~\citep{devlinBERTPretrainingDeep2019} or the improved RoBERTa~\citep{liuRoBERTaRobustlyOptimized2019} quickly adopted as baseline methods for language encoding. M-BERT~\citep{devlinBERTPretrainingDeep2019} introduced BERT's multilingual capabilities, and soon after RoBERTa and XLM-RoBERTa~\citep{conneau-etal-2020-unsupervised} followed suit. Rapid architectural advances ensued, among others, Longformer~\citep{beltagyLongformerLongDocumentTransformer2020} increased the limiting short sequence length of 512 tokens to 4096 and DeBERTaV3~\citep{he2023debertav} introduced disentangled attention that improved contextual understanding.
\cite{grecoBringingOrderRealm2024} provide a complete systematic overview of transformer models focusing on legal tasks. Most importantly, they also include an overview of non-English BERT models, such as BETO~\citep{caneteSpanishPretrainedBERT2023} and RoBERTalex~\citep{gutierrez-fandinoSpanishLegaleseLanguage2021}, and highlight the challenge of training domain- and language-specific BERT variants like the English-only Legal-BERT~\citep{chalkidis-etal-2020-legal}. 

In our work, we compare fine-tuning of smaller models like BETO~\citep{caneteSpanishPretrainedBERT2023} or XLM-RoBERTa~\citep{conneau-etal-2020-unsupervised} with the few-shot learning capabilities of larger models like GPT-4o~\citep{openaiGPT4TechnicalReport2024} or Llama-3.1-70b~\citep{dubeyLlamaHerdModels2024}. Our approach to experimentation follows \cite{mosbachFewshotFinetuningVs2023} who compare the two strategies of few-shot learning and fine-tuning while controlling for the models and the number of samples used. They find that both approaches perform similarly, but may exhibit large variations wrt. the number of model parameters and the sample counts in the training datasets or prompts. Based on these insights, we focus our evaluation on multilingual, domain-specific performance and evaluate both via fine-tuning and high few-shot sample counts.


Multilingual transfer-learning is an important technique for NLP in the legal domain, especially in the European Union~\citep{galassiUnfairClauseDetection2024}, and can be used for low-resource languages and small datasets. As an example, \cite{moroMultilanguageTransferLearning2024} proposed a transfer-learning method for extractive and abstractive summarization of legal case reports in multi- and cross-language settings, suggesting benefits of multilingual pre-training over English-only versions, a notion that our results confirm.

Domain-specific pre-training is another promising approach to transfer learning. \cite{chalkidis-etal-2020-legal} introduce Legal-BERT, a model pre-trained on English-language legal corpora. When compared with BERT on legal tasks, it manages to outperform it. 
For Spanish legal texts, \cite{gutierrez-fandinoSpanishLegaleseLanguage2021} proposed RoBERTalex. This model was trained on a large collection of Spanish legal corpora. However, its evaluation by \cite{gutierrez-fandinoMarIASpanishLanguage2022} is only on general Named Entity Recognition or classification tasks where M-BERT outperforms it significantly. 
While the optimal pre-training strategy may vary case-to-case, it is recommended to pre-train on domain-specific corpora~\citep{chalkidis-etal-2020-legal,dadasSupportSystemDetection2024}. 
However, the historic lack of domain-specific evaluation datasets~\citep{grecoBringingOrderRealm2024} made this difficult to confirm for Spanish-language legal tasks.
We investigate this by evaluating multilingual, language-specific, domain-specific, and language- and domain-specific models. We contribute the first evaluation of RoBERTalex on a Spanish-language legal classification dataset. In contrast to the general recommendation to pre-train on a domain-specific corpora~\citep{chalkidis-etal-2020-legal,dadasSupportSystemDetection2024} we actually see multilingual models like XLM-RoBERTa going head-to-head with RoBERTalex when considering the same number of trainable parameters, depending on the ambiguity of the dataset.

We even see that the more ambiguous tasks Dark Classification and Gray Classification are better addressed by few-shot learning via multilingual LLMs altogether. The training corpora of LLMs usually consist of more languages than just English, with French, German, Chinese, and Spanish being the most common minority languages~\citep{xuSurveyMultilingualLarge2024}. In variations, this applies to the models that we evaluated in our study. This helps explain the multilingual capabilities of LLMs. Coupled with cross-language transfer~\citep{moroMultilanguageTransferLearning2024}, this leads to strong generalization capabilities in the low-resource few-shot learning setting, especially with the Gray Classification dataset.

To summarize, this work contributes a broad evaluation of transformer-based language models on a novel Spanish-language legal task from Chilean law. We perform experiments with large shot counts and observe nearly linear performance increases for GPT-4o/-mini with 1-, 5- or 10-shot prompts in contrast to previous work~\citep{dominguez-olmedoLawmaPowerSpecialization2024}. In addition, and to the best of our knowledge, we are the first to evaluate RoBERTalex~\citep{gutierrez-fandinoSpanishLegaleseLanguage2021,gutierrez-fandinoMarIASpanishLanguage2022} on a domain-specific problem.

\subsection{Legal Datasets}\label{sec:rw:datasets}

Free-access legal resources are a limiting factor in the legal NLP domain~\citep{grecoBringingOrderRealm2024}. This is especially true for non-English languages and exacerbated by the focus on major jurisdictions. This section provides a concise overview of available datasets for consumer contract analysis in the English language, the multilingual, and especially the Spanish-language legal domain.

\cite{lippiCLAUDETTEAutomatedDetector2019} propose the Unfair Terms of Service dataset with 50 European consumer contracts in English annotated with 8 types of potentially abusive clauses. The authors propose models such as SVMs, Hidden-Markov Models (HMMs), Convolutional Neural Networks (CNNs), or Long-short Term Memory (LSTM) networks to detect and classify the clauses. \cite{ruggeriDetectingExplainingUnfairness2022} extend the initial dataset to a total of 100 consumer contracts, but reduce the annotations to only 5 types of potentially abusive clauses. Their proposed method provides explanations by augmenting the classifier with memory, which enriches the data available to the predictor and provides insights into the networks' reasoning. This line of investigation spawned numerous methodological developments, such as \cite{akashUnfairTOSAutomated2024} and LexGLUE~\citep{chalkidis-etal-2022-lexglue}, that perform fine-tuning studies with more recent model architectures.

The early example of a multilingual contract analysis dataset by \cite{drawzeskiCorpusMultilingualAnalysis2021} contains 25 contracts in four languages (English, German, Italian and Polish) and annotates 9 types of potentially unfair clauses from European consumer law, focusing on sentence-level annotation transfer across the four languages using ELMo~\citep{peters-etal-2018-deep} sentence embeddings coupled with Dynamic Time Warping~\citep{seninDynamicTimeWarping2008}. Recently, ~\cite{galassiUnfairClauseDetection2024} evaluated a set of methods to transfer annotations between contracts provided in different languages. They evaluate ELMo and BERT embeddings, besides the traditional Bag-of-words approach. The authors conduct their experiments on an extended multilingual Unfair ToS~\citep{drawzeskiCorpusMultilingualAnalysis2021} with 50 contracts per language (200 in total).

Several publications analyze the Terms of Services in European member states. \cite{bizzaro-etal-2024-annotation} propose a new annotation scheme for classifying relevant clauses in Italian Terms and Conditions contracts. However, their work does not include a legal analysis of whether the clauses are potentially abusive. \cite{braun-matthes-2024-agb} propose the AGB-DE dataset of 3,764 annotated clauses of German consumer contracts and compare the classification performance of a baseline SVM, fine-tuned language models and GPT-3.5. Fine-tuned models achieve higher precision, GPT-3.5 a higher recall, which is in line with our results. The authors highlight the difficulty of classification due to the ambiguity of complex clauses.
\cite{dadasSupportSystemDetection2024} publish a Polish-language dataset of 24.000 clauses labeled as safe or abusive for binary classification. Interestingly, their ML method casts the abusiveness of clauses as a problem of similarity-based information retrieval. If a clause is similar to other clauses, that were previously annotated as abusive by legal experts, it may also be abusive. Their method determines the similarity between clauses based on distances between neural embeddings, conceptually similar to work by \cite{ruggeriDetectingExplainingUnfairness2022}.


In a transfer of the methodology initially proposed by \cite{lippiCLAUDETTEAutomatedDetector2019}, but applied to Brazil, \cite{tocchiniDetectionPotentiallyNoncompliant2024} categorize the clauses of 59 Brazilian Terms of Service contracts into 10 classes with three severity levels and classify them using a SVM with Bag-of-words features.

As an example of Chinese legal tasks, we include the work of \cite{fei-etal-2024-lawbench} in our discussion. The authors propose diverse legal tasks in the Chinese language, categorized in the dimensions of memorization of knowledge, understanding of texts, and applied reasoning. They evaluate, among others, multi-label classification, and conclude that GPT-4 remains the best-performing LLM even compared with open multi-lingual and Chinese-oriented LLMs, and even LLMs fine-tuned on Chinese legal corpora.

Besides these language-specific publications, aggregate dataset collections often contribute to methodologies advances. \cite{chalkidis-etal-2022-lexglue} aggregate six English legal datasets into the Legal General Language Understanding Evaluation Evaluation (LexGLUE). \cite{10.5555/3666122.3668037} aggregate hundreds of legal datasets and evaluations in the LegalBench dataset collection.

Recently, a few Spanish-language legal datasets were published. ~\cite{aumillerStructuralTextSegmentation2021} published a large Spanish-language legal dataset of consumer Terms of Services contracts. However, their focus is on information retrieval of topics instead of potentially abusive clauses. \cite{de-gibert-bonet-etal-2022-spanish} generate four Spanish legal datasets from the MAPA project for the entity detection task. Similarly, \cite{niklausLEXTREMEMultiLingualMultiTask2023} propose the LEXTREME dataset collection that contains two datasets with Spanish law tasks from the context of the European Union, namely MultiEURLEX for document classification and MAPA for Named Entity Recognition.
To the best of our knowledge, we contribute the first dataset for Spanish-language (multi-label) clause classification, and the first legal dataset for Chilean law. 

\subsection{Limitations and Outlook}\label{sec:rw:outlook}

We see several important directions for future work. First, open questions remain as to potential improvements when combining the most advanced architectures and training methodologies, such as DeBERTaV3, with a suitable language- and domain-specific training corpora, such as those used by RoBERTalex or BETO. 
Analogously to domain-specific pre-training, \cite{dominguez-olmedoLawmaPowerSpecialization2024} propose the Lawma model by fine-tuning Llama-3 8b on 260 legal tasks, effectively emulating pre-training. Their Lawma outperforms GPT-4 in the zero- and three-shot settings. We consider both Spanish legalese pre-training and task-specific fine-tuning of LLMs with more than a billion parameters an interesting avenue for future work, as \cite{dominguez-olmedoLawmaPowerSpecialization2024} report increased performance from only a few hundred samples. The findings of \cite{blair-stanek-etal-2024-blt} on the limited reasoning capabilities of LLMs even necessitate enhancements of today's generative LLMs beyond zero-shot learning.

Similarly, recent advances in In-Context learning, such as automated reasoning or chain-of-thought~\citep{shi2023language}, may likewise lead to better scores.
The incorporation of structured knowledge into legal LLMs was recently proposed by \cite{cuiChatlawMultiAgentCollaborative2024} in their Chatlaw system and employs knowledge graphs. This method may be included in model pre-training as well as model inference. The proposed structure adheres to the established operating procedures of law firms. This approach could potentially benefit LLMs in a similar manner as it serves as a safeguard against human errors. Similarly, human workflows may be incorporated into prompt sequences~\citep{hongMetaGPTMetaProgramming2024}. First studies of such systems on the LawBench~\citep{fei-etal-2024-lawbench} dataset report a reduction in hallucinations and errors~\citep{cuiChatlawMultiAgentCollaborative2024}.

Explainable predictions are essential for acceptance by (legal) experts. \cite{schepersPredictingCitationsDutch2024} predict citations in the context of Dutch case law and their method provides explainable predictions to legal experts in the form of meta-data. They report an explicit preference for models that predict false positives, i.e., that do not miss any positive sample. 
Similarly, explaining LLMs' reasoning can be useful for estimating the predictive uncertainty of the models. LLMs that provide reasoning may even improve their predictive performance due to the auto-regressive nature of token generation~\citep{zhaoExplainabilityLargeLanguage2024}. Alternatively, similarity-based explanations that reference previously annotated samples from a database~\citep{dadasSupportSystemDetection2024} or legal explanations of a Memory Network~\citep{ruggeriDetectingExplainingUnfairness2022} may help the acceptance of fallible ML approaches in the legal field.

\section{Conclusion}\label{sec:conclusion}

A growing number of products and services are offered through online platforms. However, their Terms of Service remain enigmatic to consumers. Even if they read them, there is no alternative but to accept them. This rational apathy and information asymmetry present in modern consumer contracts may result in uncontested but potentially abusive clauses and their adverse effects on consumers.
It follows that there is a growing interest in the potential of automatic contract analysis.

We present a novel annotation scheme and a substantial dataset comprising 50 online Terms of Service. The corpus encompasses two tasks for three categories of potentially abusive clauses. Our extensive evaluation of a diverse set of transformer-based methods reveals that multiple factors impact the detection and classification rates, including language- and domain-specific pre-training, the number of samples in few-shot prompts, and architectural considerations.
Potentially abusive clauses are detected with macro-F1 scores ranging from between $79\%$ to $89\%$ and micro-F1 scores of $95\%$ to $96\%$. The highest classification rates range from between $60\%$ to $70\%$ macro-F1 score and $64\%$ to $80\%$ micro-F1 score. This underscores the wide range of difficulties inherent in the novel datasets.

This work introduces the first Spanish-language classification dataset of legal clauses. Moreover, it applies law from outside a major jurisdiction, specifically the Chilean legal code.
We also present the first comprehensive study on Spanish-language and multilingual transformer-based models on legal classification tasks in Spanish.
This paper serves as the foundation for substantial future research and the implementation of its methodology to benefit Chilean and Latin American consumers.


\backmatter







\begin{appendices}

\section{Additional Dataset Information}\label{secA1}

This appendix provides supplementary information on dataset statistics, see Table~\ref{tab:stats_total}

\begin{table}
\begin{tabular}{ |p{2cm}|p{2cm}|p{2cm}| }
\hline
\multicolumn{3}{|c|}{Dataset statistics} \\
\hline
\textbf{Code} & \textbf{Absolute Number} & \textbf{Percentage} \\ \hline
ok                 & 4165 & 0.770868 \\ \hline
na & 284  & 0.052563 \\ \hline
ltd                & 138  & 0.025541 \\ \hline
lpc pro& 133  & 0.024616 \\ \hline
cr                 & 91   & 0.016842 \\ \hline
des risk           & 64   & 0.011845 \\ \hline
cc rc & 54   & 0.009994 \\ \hline
des reser          & 48   & 0.008884 \\ \hline
bfe                & 45   & 0.008329 \\ \hline
lpc & 44   & 0.008144 \\ \hline
des uni            & 43   & 0.007959 \\ \hline
lpc int & 40   & 0.007403 \\ \hline
lpc jus & 36   & 0.006663 \\ \hline
nod                & 35   & 0.006478 \\ \hline
des def            & 33   & 0.006108 \\ \hline
des det            & 30   & 0.005552 \\ \hline
er                 & 27   & 0.004997 \\ \hline
des inf            & 24   & 0.004442 \\ \hline
des lic            & 14   & 0.002591 \\ \hline
ch                 & 10   & 0.001851 \\ \hline
ter                & 8    & 0.001481 \\ \hline
\end{tabular}
\caption{Dataset statistics.}
\label{tab:stats_total}
\end{table}

\section{Cost Considerations}\label{secA2}

The availability of open LLMs such as Llama3.1 or Qwen2.5 presents serious competition to commercial services like the hosted GPT-4o (-mini) models. However, the price of $1.25$ USD for 1M input tokens of OpenAI's batch API may be considered relatively low, due to a rapid fall in prices compared to legacy models such as GPT-4. Hence, the fundamental choice between commercial or open LLMs depends not only on the expected scale of the amount of data to process or the size of the datasets, but also on data protection considerations, that are crucial for sensitive legal data.

\end{appendices}

\bibliography{bibtex}

\end{document}